\documentclass{article}

\usepackage[preprint, nonatbib]{neurips_2026}
\usepackage[sort&compress,numbers]{natbib}

\usepackage[utf8]{inputenc} 
\usepackage[T1]{fontenc}    
\usepackage{hyperref}       
\usepackage{url}            
\usepackage{booktabs}       
\usepackage{amsfonts}       
\usepackage{nicefrac}       
\usepackage{microtype}      
\usepackage{xcolor}         
\usepackage{tikz}
\usepackage{graphicx}
\usepackage{pifont}  
\usepackage{multirow}  
\usetikzlibrary{positioning, arrows.meta, calc, fit, backgrounds, shapes.geometric, decorations.pathreplacing}
\usepackage{amsmath}
\usepackage{xspace}
\usepackage{subcaption}
\usepackage{enumitem}

\makeatletter
\DeclareRobustCommand\onedot{\futurelet\@let@token\@onedot}
\def\@onedot{\ifx\@let@token.\else.\null\fi\xspace}

\def\eg{\emph{e.g}\onedot}

\makeatother

\title{Reinforcing VLAs in Task-Agnostic World Models}

\author{%
  Yucen Wang$^{2}$\thanks{Interns at Microsoft.} ~~~~~ Rui Yu$^{3}$ ~~~~~ Fengming Zhang$^{2}$ ~~~~~ Junjie Lu$^{5}$ ~~~~~ \textbf{Xinyao Qin}$^{6}$ \\ ~~~~~ \textbf{Tianxiang Zhang}$^{4}$ ~~~~~ \textbf{Kaixin Wang}$^{1}$  ~~~~~ \textbf{Li Zhao}$^{1}$  \\\\
  $^1$ Microsoft Research Asia \quad $^2$ Nanjing University \\
  $^3$ University of Illinois at Urbana-Champaign \quad $^4$ Wuhan University \\
  $^5$ University of Technology Sydney \quad $^6$ Tsinghua University
}

\begin{document}

\maketitle

\begin{abstract}
Post-training Vision-Language-Action (VLA) models via reinforcement learning (RL) in learned world models has emerged as an effective strategy to adapt to new tasks without costly real-world interactions.
However, while using imagined trajectories reduces the sample complexity of policy training, existing methods still heavily rely on task-specific data to fine-tune both the world and reward models, fundamentally limiting their scalability to unseen tasks.
To overcome this, we argue that world and reward models should capture transferable physical priors that enable zero-shot inference.
We propose \textbf{RAW-Dream} (\textbf{R}einforcing VLAs in task-\textbf{A}gnostic \textbf{W}orld \textbf{Dream}s), a new paradigm that completely disentangles world model learning from downstream task dependencies.
RAW-Dream utilizes a world model pre-trained on diverse task-free behaviors for predicting future rollouts, and an off-the-shelf Vision-Language Model (VLM) for reward generation.
Because both components are task-agnostic, VLAs can be readily finetuned for any new task entirely within this zero-shot imagination.
Furthermore, to mitigate world model hallucinations, we introduce a dual-noise verification mechanism to filter out unreliable rollouts.
Extensive experiments across simulation and real-world settings demonstrate consistent performance gains, proving that generalized physical priors can effectively substitute for costly task-dependent data, offering a highly scalable roadmap for VLA adaptation.
\end{abstract}

\section{Introduction}
\label{sec:intro}

Vision-Language-Action (VLA) policies trained via imitation learning have advanced robotic manipulation considerably
~\cite{black2024pi0, intelligence2025pi, kim2024openvla, chen2025villa}, yet remain brittle beyond their training distribution.
Reinforcement learning (RL) can in principle address this through trial-and-error improvement, but online RL on physical robots is prohibitively costly.
A rapidly growing line of work resolves this tension by leveraging action-conditioned video world models (WMs) as virtual simulators, in which VLA policies are refined entirely through imagination
~\cite{zhu2025wmpo, jiang2026wovr, li2025vla, liu2026world, xiao2025world, sharma2026world, guo2026vlaw, yang2026rise, zhang2025reinforcing, hung2025nora, zhang2026towards}, 
yielding substantial gains over pure imitation baselines.

However, a critical limitation remains: \textbf{existing methods operate exclusively on tasks that their WMs were already trained on}.
While real-world deployment routinely demands rapid adaptation to novel tasks (\eg, new objects, rearranged layouts, unfamiliar instructions), prior methods rely on substantial rollout data collected specifically on the target task to train their WMs~\cite{zhu2025wmpo, jiang2026wovr, guo2026vlaw, liu2026world, sharma2026world}.
Reward mechanisms are similarly entangled, requiring task-specific classifiers or reference trajectories\cite{zhu2025wmpo, jiang2026wovr, li2025vla, liu2026world}.
The primary appeal of using a WM is to bypass costly real-world interactions and act as a general-purpose simulator; yet, current methods require knowing the target tasks in advance and collecting extensive real-world data before the pipeline can even begin.
Consequently, the WM must be rebuilt whenever new tasks arise, reducing it to a costly, single-use data-augmentation tool and structurally precluding unseen-task generalization.

\begin{figure}[t]
    \centering
    \includegraphics[width=\textwidth]{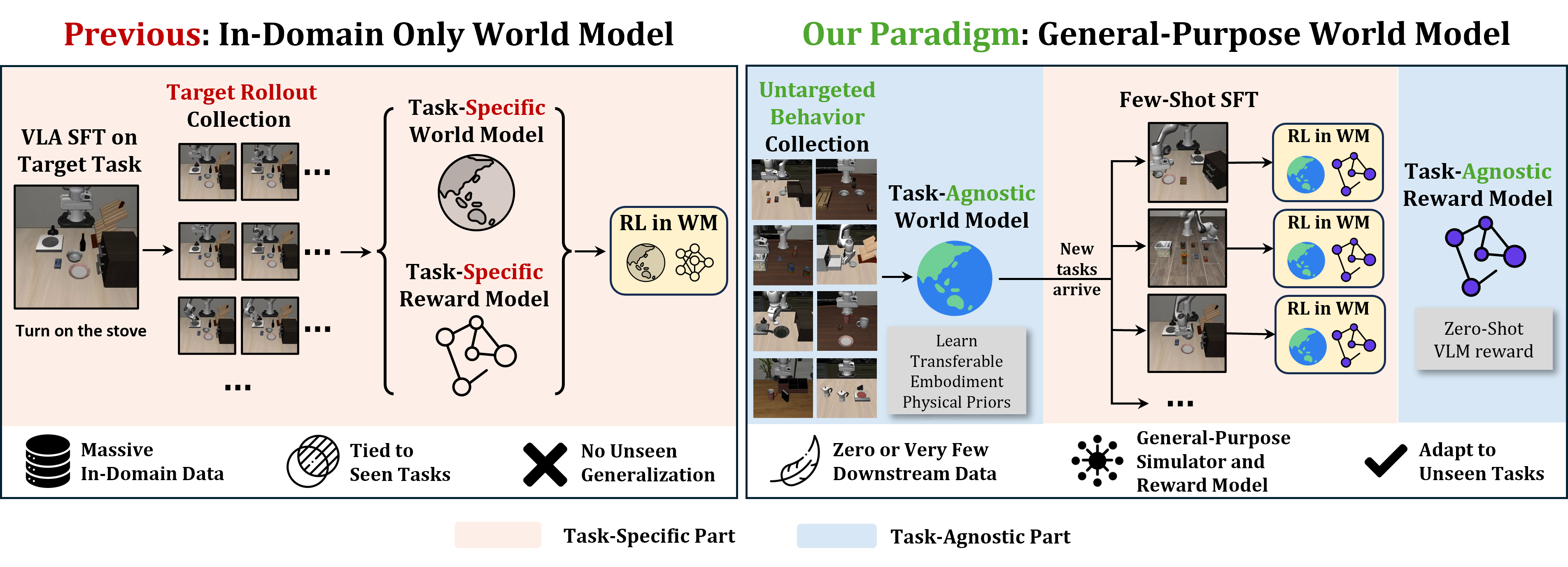}
    \caption{
    \textit{Left:} Previous WM-based RL pipelines for VLA post-training tightly couple the WM and reward models to known target tasks, requiring thousands of in-domain rollouts, precluding unseen adaptation.
    \textit{Right:} RAW-Dream decouples dynamics learning from task semantics. A general-purpose WM pre-trained on diverse task-free behaviors captures transferable physical priors, while a foundation VLM provides zero-shot rewards. Together, they enable data-efficient RL on novel tasks.
    }
    \label{fig:motivation}
\vskip -10pt
\end{figure}

To break this task-specific data bottleneck, we draw on a fundamental observation: the physical dynamics of a robotic workspace are inherently task-independent.
A bowl slides identically regardless of whether the instruction is ``put the bowl on the shelf'' or ``push the bowl aside''.
Following this insight, we propose \textbf{RAW-Dream} (\textbf{R}einforcing VLAs in task-\textbf{A}gnostic \textbf{W}orld \textbf{Dream}s), a new paradigm for VLA post-training where both the WM and the reward function are strictly \emph{task-agnostic}.
To achieve this, the WM is pre-trained on diverse, broad-coverage interaction data (\eg, play data) to serve as a reusable, general-purpose physical simulator~\cite{yin2026playworld, chandra2025diwa, mazzaglia2024genrl, wang2025founder, sekar2020planning, lu2022challenges}.
Concurrently, we leverage the generalized reasoning of off-the-shelf vision-language models (VLMs)~\cite{sharma2026world, zhang2025reinforcing}, which can provide reliable, zero-shot reward signals by judging task success directly from the imagined video rollout and text instruction.

Under RAW-Dream, adapting to novel downstream tasks becomes highly data-efficient.
The VLA policy is first anchored to the new task semantics via lightweight supervised fine-tuning (SFT) on minimal expert demonstrations.
It is then optimized via RL entirely inside the frozen, task-agnostic WM using the zero-shot VLM reward.
We argue this represents a significantly more scalable paradigm for WM-based RL: the simulator and reward function are prepared once and generalize broadly, leaving only the policy to adapt to new instructions.
Consequently, the target-task rollout data requirement for the WM and reward drops from thousands of trajectories to \textbf{zero}.
Even if optional few-shot WM fine-tuning is desired to bridge visual domain gaps, the total data budget remains vastly below that of training from scratch.

We build our action-conditioned WM upon the WAN~2.1 video generation model~\cite{wan2025wan} and adopt OpenVLA-OFT~\cite{kim2025fine, li2025simplevla, zhu2025wmpo, jiang2026wovr} as the policy.
We optimize the VLA via GRPO~\cite{shao2024deepseekmath}, guided exclusively by binary, zero-shot success rewards from a frozen Qwen3-VL model~\cite{bai2025qwen3}.
While we build upon established architectures, integrating them within our task-agnostic paradigm effectively eliminates target-task rollout constraints, paving the way for highly efficient downstream adaptation.
Finally, because querying a general-purpose WM on unseen tasks naturally increases the risk of hallucinated successes, we introduce a lightweight \emph{dual-noise verification} mechanism that requires the VLM reward to remain consistent across rollouts generated under different initial diffusion noises.
This effectively filters out hallucinations and curbs false-positive reward hacking.

We validate our paradigm in simulation and real-world settings where WMs are built strictly from broad, target-task-independent data.
On the LIBERO benchmark~\cite{liu2023libero}, we train the WM exclusively on LIBERO-90 data and optimize the VLA across four held-out task suites (Spatial, Object, Goal, Long) entirely unseen during WM training.
Our approach lifts the performance of a 1-shot SFT baseline by \textbf{+8.9\%} and outperforms online RL that consumes $50\times$ more real-world data.
On physical robots, we adapt policies to novel manipulation tasks using a WM pre-trained solely on diverse, uncurated play data.
Starting from a 3-shot SFT baseline per task, RAW-Dream delivers a \textbf{+21.7\%} absolute improvement in success rate.
Our results validate that broad physical priors can effectively substitute for costly in-domain data, pointing to a clean and actionable scaling roadmap for WM-based VLA RL optimization.

\section{Related work}
\label{sec:related}

\textbf{Finetuning VLA with reinforcement learning} \quad
Finetuning is a crucial step for adapting pre-trained Vision-Language-Action (VLA) models to unseen tasks.
Inspired by advancements in reinforcement learning (RL) for large language models, there is growing interest in applying RL to finetune VLAs.
We categorize existing approaches based on the source of their RL interaction data.
One line of work (\eg, VLA-RL~\cite{lu2025vla}, SimpleVLA-RL~\cite{li2025simplevla}, $\pi_{\texttt{RL}}$~\cite{chen202500_0texttt0rl000}) uses simulated environments for data collection.
However, these methods are often bottlenecked by the sim-to-real gap and the substantial engineering effort required to build high-fidelity digital twins.
Alternatively, methods such as RECAP~\cite{intelligence2025000000_0006000} and RLT~\cite{xu2026rl} collect data directly from physical robots, effectively bypassing sim-to-real challenges but suffering from sample inefficiency.
In contrast, our approach aligns with a recent paradigm that generates interaction data using a learned world model (which we review in detail in the next subsection).
Driven by the strong predictive capabilities of modern world models to synthesize realistic future rollouts~\cite{team2025evaluating, quevedo2025worldgym0, tseng2025scalable}, performing RL finetuning within a world model enjoys high sample efficiency while significantly mitigating the sim-to-real gap.

\textbf{Learning world models for VLA finetuning}\quad
Closely related to our work is a recent line of research that finetunes VLAs with the help of learned world models.
These works typically use a world model to generate synthetic rollouts and apply RL on these imagined trajectories to optimize the VLA policy without physical execution.
Some frameworks~\cite{zhu2025wmpo, jiang2026wovr, sharma2026world, zhang2026towards, xiao2025world} utilize world models as static simulators for post-training policies, while others~\cite{guo2026vlaw, liu2026world, yang2026rise} explore closed-loop systems where the policy and world model are iteratively co-trained.
To provide reward signals for these imagined rollouts, existing literature generally relies on specifically engineered~\cite{li2025vla} or task-finetuned~\cite{zhu2025wmpo, jiang2026wovr, guo2026vlaw}) reward functions, though a few methods~\cite{hung2025nora, zhang2025reinforcing,sharma2026world} explored zero-shot VLM rewards.
However, a fundamental limitation shared across these approaches is their reliance on task-dependent world models.
Unlike these approaches, our work takes a step towards a more generalizable framework by pairing a task-agnostic world model, capable of simulating unseen tasks out-of-the-box, with off-the-shelf VLM rewards, entirely eliminating the need for task-specific finetuning on both fronts.
We believe this paradigm is inherently more scalable and can be readily upgraded as more powerful foundational world models and VLMs become available.

\section{Method}
\label{sec:method}

Unlike prior pipelines that directly optimize for known target tasks, RAW-Dream operates under the strict constraint of target-task-agnostic VLA-RL optimization in WMs. 
This forces us to specially design the WM training / inference process 
and the RL optimization loop in the WM to handle unseen generalization.
RAW-Dream focuses on two core components:
\textbf{(i)}~An action-conditioned video world model that learns transferable physical dynamics from the embodiment's diverse behaviors, 
enabling reliable rollouts even under severe target-domain data constraints.
\textbf{(ii)}~An imagination-based RL pipeline driven by a zero-shot VLM reward in GRPO, 
equipped with a dual-noise verification mechanism to filter the severe WM hallucinations caused by target-data scarcity.

\subsection{World Modeling under Target-Domain Data Scarcity}
\label{sec:method_wm}

\textbf{Establishing physical priors.} 
Under strict data constraints for target domains, 
the WM must learn generalizable physical dynamics from task-agnostic data. 
We train our WM on a broad mixture of diverse task-free behaviors 
(\eg, general uncurated play data or noisy exploratory rollouts collected by executing VLA policies), 
covering varied success and failure patterns across a wide range of physical interactions and transitions, 
forcing the model to learn versatile physical dynamics rather than overfitting to the semantics of any single task or expert demonstration.
More details can be found in our experiment sections.

\textbf{Architecture and action-conditioning.} 
We build our WM on the pre-trained Wan~2.1-T2V-1.3B Diffusion Transformer (DiT) video generation model backbone, 
operating within the VAE latent space. 
We finetune it into an action-conditioned generator via adaptive layer normalization (AdaLN) \cite{peebles2023scalable}.
The action embeddings are projected into action-specific scale, shift, and gate parameters by a from-scratch MLP, 
which are then fused with the original diffusion timestep AdaLN parameters,
and modulate the attention results in each DiT block. 
This mechanism provides control signals to condition the denoising process on the actions \cite{zhu2025irasim, he2026pre, wang2025co, ali2025world}.
To prevent generation from future-action leakage, 
we strictly enforce causal masking for temporal attention operation in the DiT blocks. 
The network $v_\theta$ is supervised using a unified rectified flow matching objective~\cite{liu2022flow} to predict the continuous velocity field $\mathbf{v}^{*} = \boldsymbol{\epsilon} - \mathbf{z}^{0}$:
\begin{equation}
    \mathcal{L} = \mathbb{E}_{\tau,\,\mathbf{z},\,\boldsymbol{\epsilon},\, \mathbf{a}} \left[ \big\lVert v_\theta(\mathbf{z}^{\tau}, \tau, \mathbf{a}, \mathbf{z}_{ctx}) - \mathbf{v}^{*} \big\rVert^2 \right],
\end{equation}
where $\tau$ is the diffusion timestep, $\mathbf{z}^{\tau}=\tau \boldsymbol{\epsilon} + (1-\tau) \mathbf{z}^0 $ is the noisy latent state, 
$\mathbf{z}^0$ is the ground-truth latent state,
$\boldsymbol{\epsilon} \sim \mathcal{N}(\mathbf{0}, \mathbf{I})$ is the noise, 
$\mathbf{a}$ denotes the conditioned actions, 
and $\mathbf{z}_{ctx}$ represents the context latent states.

\textbf{Long-horizon autoregressive rollout.} 
To make long-horizon prediction, 
the WM generates video autoregressively, conditioned on recently generated states as context. 
To accommodate the policy's action chunk size (8 in OpenVLA-OFT), the WM predicts 2 states per step in inference. 
We employ a large context window of 6 for reliable generation.
During training, context frames are corrupted with diffusion noise (timesteps sampled from $[0, 300]$), 
compelling the model to make predictions from imperfect contexts. 
Also, we apply conditioning masks to construct context and generated states of variable lengths. 
In inference, we set a fixed noise timestep $\tau=50$ for the context frames.

\textbf{First-frame anchoring and progressive noise.} 
When rolling out in entirely unseen target domains, 
the initial observation $\mathbf{o}_{0}$ is the only ground-truth visual reference. 
We condition every autoregressive step on this first frame to ensure scene consistency, using 7 context and predicting 2 latent states per step. 
However, this also induces \emph{first-frame ghosting} in zero-shot WM transfer: 
lacking domain familiarity, the model over-anchors to the reliable $\mathbf{o}_{0}$, 
stubbornly pasting initial object layouts into later autoregressive chunks, 
creating ``ghosts'' of the first frame that corrupt both dynamics and VLM reward judgments in RL. 
To resolve this, we introduce \emph{progressive anchor noise} at inference. 
We gradually increase diffusion timesteps on the anchor as the autoregressive rollout progresses. 
This smoothly degrades the anchor's influence over time, 
forcing the model to rely on recent context while preserving $\mathbf{o}_{0}$'s role in maintaining scene consistency. 
We provide further WM implementation details in Appendix~\ref{sec:appendix_wm_details} and qualitative results of first-frame ghosting in Appendix~\ref{sec:appendix_ghost}.

\subsection{Policy Optimization in Task-Agnostic World Model}
\label{sec:method_rl}

Starting from an initial visual observation and a task instruction, 
the OpenVLA-OFT policy samples actions to interact with the learned WM autoregressively, 
generating a group of imagined trajectories $\{\tau_i\}_{i=1}^{G}$. 
Since our paradigm strictly precludes collecting target-task rollouts to train success classifiers, 
we evaluate these imagined videos using an off-the-shelf VLM (Qwen3-VL), 
yielding a binary outcome $R_i$ for each trajectory $\tau_i$. 
This group is then directly used to update the policy via GRPO.

While this provides a stable optimization pipeline, it inevitably incurs hallucinations on unseen tasks. 
The policy may discover visually plausible but dynamically unfaithful rollouts that fool the VLM to give false positive rewards, a risk amplified by our task-agnostic WM's lack of downstream training data. To address this, we draw inspiration from model-based offline RL that penalizes the rewards of unreliable transitions based on prediction uncertainty \cite{yu2020mopo, kidambi2020morel}, 
and present a tractable analog tailored for modern DiT-based WMs and RL based on binary outcome rewards. 
Because our WM employs long-horizon autoregressive generation via diffusion denoising processes, 
the randomness and uncertainty naturally compounds over steps, 
especially when generalizing to unseen tasks and scenes. 
Therefore, rollouts regenerated under different initial diffusion noise are more likely to yield divergent outcomes under out-of-distribution dynamics
compared to reliable transitions.

Building on this, we introduce \textbf{Dual-Noise Verification} (DNV) to detect and penalize hallucinations. 
For any imagined trajectory $\tau_i$ within the GRPO group that was marked successful by the VLM, 
we replay its action sequence $\{a_{i,t}\}_{t=1}^{T_i}$ in the WM using independently resampled initial diffusion noise at each autoregressive step.
If the VLM evaluates this newly generated video as a failure, the initial success is flagged as unfaithful and the rollout is discarded from the group.
Crucially, we detect hallucinations directly through the disagreement of the VLM reward rather than video generation discrepancies. 
Low-level video metrics are often unaligned with task semantics; 
by contrast, utilizing the VLM's judgment provides a semantic validation that directly grounds the final reward penalty.

In our GRPO objective based on DNV, 
we compute advantages using group-relative returns over the reliable subset $\mathcal{R}$ of size $G_{\mathcal{R}}$. More details and analysis on DNV can be found in our Appendix.
\begin{equation}
\mathcal{J}(\theta)
= \mathbb{E}_{s_0 \sim \mathcal{D}, \{\tau_i\} \sim \pi_{\theta_{\text{old}}}}
\left[\frac{1}{G_{\mathcal{R}}} \sum_{i \in \mathcal{R}} \frac{1}{T_i} \sum_{t=1}^{T_i}\min\big(r_{i,t}(\theta)\hat{A}_i,\;
\mathrm{clip}(r_{i,t}(\theta), 1-\epsilon_{\text{low}}, 1+\epsilon_{\text{high}})\hat{A}_i\big)\right],
\end{equation}
with
\begin{equation}
r_{i,t}(\theta)
= \frac{\pi_{\theta}(a_{i,t} \mid s_{i,t})}{\pi_{\theta_{\text{old}}}(a_{i,t} \mid s_{i,t})},
\qquad
\hat{A}_i
= \frac{R_i - \mathrm{mean}(\{R_j\}_{j \in \mathcal{R}})}{\mathrm{std}(\{R_j\}_{j \in \mathcal{R}})}.
\end{equation}

Here $r_{i,t}(\theta)$ is the probability ratio between new and old policies at step $t$ of trajectory $\tau_i$, 
$R_i = R(\tau_i)$, and $\hat{A}_i$ is the normalized group-relative advantage over the reliable subset.

\section{Experiments}
\label{sec:experiments}

We conduct extensive experiments across both the LIBERO simulation benchmark~\cite{liu2023libero} and our real-world robotic setups to systematically validate the effectiveness of RAW-Dream. 
Our experiments are designed to answer the following four questions:
\begin{itemize}[leftmargin=15pt]
    \item \textbf{Q1:} Can an action-conditioned video world model trained on broad, downstream-task-agnostic trajectory data effectively predict and simulate unseen target tasks? (Section~\ref{sec:exp_wm})
    \item \textbf{Q2:} Can RL within the task-agnostic world model using task-agnostic VLM reward effectively improve VLA performance under minimal task-specific data constraints? (Section~\ref{sec:exp_rl})
    \item \textbf{Q3:} How do the zero-shot VLM reward and the dual-noise verification mechanism contribute to the success of imagination-based RL? (Section~\ref{sec:exp_ablation})
    \item \textbf{Q4:} Does this paradigm successfully transfer to physical robots to facilitate real-world manipulation tasks? (Section~\ref{sec:exp_real_world})
\end{itemize}

\subsection{Experimental settings for simulation}
\label{sec:exp_setup}

We instantiate our paradigm on the LIBERO benchmark using a strict evaluation protocol to mirror our proposed downstream-agnostic paradigm: 
constructing a foundational WM from LIBERO-90, 
and subsequently evaluating its simulation fidelity and RL performance on four entirely unseen downstream task suites under minimal target-task data constraints. 

\textbf{A general-purpose WM from LIBERO-90.} The \textbf{LIBERO-90} benchmark (90 tasks, containing $\sim$4{,}500 expert demonstrations) offers a rich diversity of scenes and manipulation tasks. 
We leverage this benchmark as a foundational source to construct a general-purpose WM for the LIBERO embodiment. 
Specifically, we first train an OpenVLA-OFT policy via multi-task SFT on the curated LIBERO-90 expert demonstrations. 
To build a comprehensive physical prior,
we collect broad exploration data by executing this SFT policy with injected Gaussian noise on its action outputs across all 90 tasks.
This yields a large,
diverse corpus of rollouts spanning successes and failures across a broad range of tasks with diverses,
yet targeting no specific downstream task,
which can be seen as the \emph{proxy play data} for LIBERO.
The WM is then trained entirely on these rollouts,
forcing the model to internalize versatile, broadly applicable physical dynamics.

\textbf{Unseen downstream evaluation.} We then employ four highly distinct held-out downstream task suites for evaluation: 
\textbf{Spatial} (novel item arrangements), \textbf{Object} (novel objects), \textbf{Goal} (novel semantic goals), and \textbf{Long} (long-horizon sequences). 
Each suite comprises 10 distinct tasks. 
Crucially, the WM has never observed any data from these four suites during pretraining on LIBERO-90. 
When confronted with a novel target suite, we follow a two-phase VLA policy learning pipeline. 
First, via semantic anchoring (multi-task \textbf{1-shot SFT}), 
the VLA policy previously trained on LIBERO-90 is further adapted using exactly \emph{one} expert demonstration per task (10 demonstrations per suite), 
anchoring novel task semantics with minimal demonstration data. 
Second, via multi-task \textbf{RL in WM imagination}, 
the anchored policy is improved through trial-and-error entirely within the WM via GRPO,
using the zero-shot binary success reward provided by the frozen Qwen3-VL~\cite{bai2025qwen3}.

\textbf{World model adaptation conditions.}
Since the RL phase is conducted entirely within the WM, 
the simulator's fidelity directly governs the policy's improvement ceiling. 
To systematically probe this dependency, 
we evaluate the WM fidelity and the above WM-based RL pipeline under four conditions that span a wide spectrum of target-domain data exposure:
\begin{itemize}
    \item \textbf{Zero-Shot WM:} The foundational WM trained on LIBERO-90 is fully frozen, 
    observing exactly zero target-suite rollouts.
    \item \textbf{Co-Train WM:} 
    Since the zero-shot WM may lack visual familiarity with novel target scenes,
    we anchor it to the target domain at minimal cost by mixing the 10 expert demonstrations from 1-shot SFT into the LIBERO-90 WM training data and jointly fine-tuning,
    avoiding overfitting on scarce expert-only data while incurring zero additional collection budget.
    \item \textbf{ID-FT WM (In-Domain FineTuning):} The 1-shot SFT VLA policy is evaluated on the target suite, 
    yielding 500 rollouts (50 per task) that are used to fine-tune the foundational LIBERO-90 WM. 
    This serves as an upper bound for our architecture.
    \item \textbf{WoVR WM~\cite{jiang2026wovr}:} A recent state-of-the-art method that trains a 5B, WAN2.2-based WM from scratch on 2,500 per-target-task-suite rollouts without any broad pre-training, 
    providing a direct comparison against data-intensive, task-specific WM construction.
\end{itemize}

\textbf{Evaluation.} For WM fidelity, we evaluate the action-conditioned video generation quality of each WM on a fixed validation dataset for each task suite (each containing 500 trajectories unseen in training),
using standard metrics: PSNR, SSIM, LPIPS and FVD. LPIPS and SSIM scores are scaled ×100 for compact display. 
For VLA policy evaluation in SFT and RL, 
each method is evaluated by executing the learned policy in the real LIBERO simulator for 50 episodes per task, 
reporting per-suite average success rate (\%). Further evaluation details can be found in our Appendix~\ref{sec:appendix_impl}.

\begin{table}[t]
  \caption{\textbf{Action-conditioned video prediction quality on unseen task suites using different WMs.} 
  ``Tgt.\ data'' denotes the number of target-task-suite trajectories used for WM training.
  Average trajectory length of validation datasets in parentheses lies below each suite name.}
  \label{tab:wm_quality}
  \centering
  \small
  \setlength{\tabcolsep}{4pt}
  \begin{tabular}{cl c cccc}
    \toprule
    Suite & WM Variant & Tgt.\ data & PSNR$\uparrow$ & SSIM$\uparrow$ & LPIPS$\downarrow$ & FVD$\downarrow$ \\
    \midrule
    \multirow{4}{*}{\shortstack{Spatial\\{\scriptsize(Avg Length: 164)}}}
    & Zero-Shot WM           & 0   & 19.17 & 80.24 & 12.51 & 92.29 \\
    & Co-Train WM     & 10  & 21.34 & 84.58 & 8.26 & 60.76 \\
     & \textbf{ID-FT WM}    & 500 & \textbf{24.99} & \textbf{89.58} & \textbf{4.20} & \textbf{23.52} \\
    & WoVR WM            & 2500 & 22.80 & 86.95 & 6.61 & 45.39 \\
    \midrule
    \multirow{4}{*}{\shortstack{Object\\{\scriptsize(Avg Length: 208)}}}
    & Zero-Shot WM           & 0   & 19.36 & 76.59 & 13.85 & 233.29 \\
    & Co-Train WM    & 10  & 19.94 & 82.23 & 9.77 & 114.06 \\
        & \textbf{ID-FT WM}     & 500 & \textbf{25.91} & \textbf{90.28} & \textbf{3.53} & \textbf{26.82} \\
    & WoVR WM            & 2500 & 22.73 & 86.59 & 6.15 & 92.11 \\
    \midrule
    \multirow{4}{*}{\shortstack{Goal\\{\scriptsize(Avg Length: 212)}}}
    & Zero-Shot WM           & 0   & 19.50 & 83.11 & 11.19 & 89.83 \\
    & Co-Train WM     & 10  & 21.94 & 87.21 & 7.17 & 50.33 \\
        & \textbf{ID-FT WM}      & 500 & \textbf{25.33} & \textbf{91.15} & \textbf{3.79} & \textbf{21.65} \\
    & WoVR WM            & 2500 & 22.92 & 88.67 & 6.40 & 41.78 \\
    \midrule
    \multirow{4}{*}{\shortstack{Long\\{\scriptsize(Avg Length: 472)}}}
    & Zero-Shot WM           & 0   & 19.14 & 82.87 & 12.96 & 56.43 \\
    & Co-Train WM     & 10  & 19.60 & 84.00 & 11.15 & 44.48 \\
        & \textbf{ID-FT WM}      & 500 & \textbf{20.32} & \textbf{84.85} & \textbf{8.96} & \textbf{38.84} \\
    & WoVR WM            & 2500 & 18.03 & 82.04 & 12.97 & 80.52 \\
    \bottomrule
  \end{tabular}
\end{table}

\subsection{World model simulation quality}
\label{sec:exp_wm}

Table~\ref{tab:wm_quality} compares four WM variants across all four unseen target suites.
Under the strict zero-shot setting, apart from LIBERO-Object 
which introduces entirely novel objects that severely shock the input distribution (FVD=233), 
the WM maintains a remarkably capable baseline across other suites. 
This demonstrates that physical priors learned from broad, downstream-agnostic data can effectively generalize to unseen domains.
Incorporating the exact same 10 demonstrations used for SFT semantic anchoring into the WM (Co-Train), 
which demands zero additional data collection budget, yields a stark enhancement in fidelity,
rapidly approximating the performance of WoVR (particularly on Spatial and Goal), which relies on massive in-domain data.

Furthermore, on LIBERO-Long, the closest suite to LIBERO-90, the 
Zero-Shot WM already eclipses WoVR despite WoVR having access to massive in-domain data. 
This highlights that when underlying scenes overlap with the pre-training distribution, 
zero-shot transfer is highly effective, 
and validates the superiority of our WM architecture and inference machanism in long-horizon simulation.

The most critical takeaway arises when comparing our ID-FT upper-bound against WoVR. 
By injecting just 500 target rollouts into our pre-trained foundation, 
the WM outperforms WoVR (trained from scratch on 2,500 target rollouts) across all suites and metrics, 
confirming that broad, downstream-independent physical priors outweigh brute-force data stacking within a single target domain.

\begin{table}[t]
  \caption{\textbf{Success rate (\%) on four unseen LIBERO suites.}
  The VLA is first trained on LIBERO-90, then adapted to each suite using 1-shot SFT.
  Bold rows correspond to our method with different WM conditions.
  Bold rows correspond to our method with different WM conditions.
  ``Tgt.\ data'' refers to target-suite real rollout budget used in the whole policy learning process, including data for WM and VLA fine-tuning. 
  Deltas in parentheses relative to 1-shot SFT.}
  \label{tab:main_rl}
  \centering
  \small
  \setlength{\tabcolsep}{4pt}
  \renewcommand{\arraystretch}{1.15}
  \resizebox{\textwidth}{!}{%
  \begin{tabular}{@{}l l c ccccc@{}}
    \toprule
    Category & Method & Tgt.\ data & Spatial & Object & Goal & Long & Avg. \\
    \midrule
    \multirow{2}{*}{No RL}
    & Zero-Shot from 90  & 0 & 3.4 & 0.0 & 4.2 & 7.0 & 3.7 \\
    & 1-shot SFT & 10 & 54.6 & 46.4 & 52.2 & 20.2 & 43.4 \\
    \midrule
    \multirow{2}{*}{\shortstack[l]{RL in simulator\\(GT reward)}}
    & Online RL (Short) & 522 & 58.4\,{\scriptsize(+3.8)} & 60.2\,{\scriptsize(+13.8)} & 55.2\,{\scriptsize(+3.0)} & 17.6\,{\scriptsize(-2.6)} & 47.9\,{\scriptsize(+4.5)} \\
    & Online RL (Long) & 2570 & 68.8\,{\scriptsize(+14.2)} & 78.8\,{\scriptsize(+32.4)} & 65.2\,{\scriptsize(+13.0)} & 22.4\,{\scriptsize(+2.2)} & 58.8\,{\scriptsize(+15.4)} \\
    \midrule
    \multirow{4}{*}{\shortstack[l]{RL in WM\\(VLM reward)}}
    & \textbf{Zero-Shot WM} & \textbf{10} & \textbf{65.8}\,{\scriptsize\textbf{(+11.2)}} & \textbf{47.2}\,{\scriptsize\textbf{(+0.8)}} & \textbf{60.2}\,{\scriptsize\textbf{(+8.0)}} & \textbf{35.8}\,{\scriptsize\textbf{(+15.6)}} & \textbf{52.3}\,{\scriptsize\textbf{(+8.9)}} \\
    & \textbf{Co-Train WM} & \textbf{10} & \textbf{73.2}\,{\scriptsize\textbf{(+18.6)}} & \textbf{60.2}\,{\scriptsize\textbf{(+13.8)}} & \textbf{58.4}\,{\scriptsize\textbf{(+6.2)}} & \textbf{36.6}\,{\scriptsize\textbf{(+16.4)}} & \textbf{57.1}\,{\scriptsize\textbf{(+13.7)}} \\
    & \textbf{ID-FT WM} & \textbf{510} & \textbf{82.0}\,{\scriptsize\textbf{(+27.4)}} & \textbf{79.8}\,{\scriptsize\textbf{(+33.4)}} & \textbf{63.4}\,{\scriptsize\textbf{(+11.2)}} & \textbf{38.6}\,{\scriptsize\textbf{(+18.4)}} & \textbf{66.0}\,{\scriptsize\textbf{(+22.6)}} \\
    & WoVR WM & 2510 & 81.6\,{\scriptsize(+27.0)} & 71.6\,{\scriptsize(+25.2)} & 66.4\,{\scriptsize(+14.2)} & 24.0\,{\scriptsize(+3.8)} & 60.9\,{\scriptsize(+17.5)} \\
    \bottomrule
  \end{tabular}
  }
\end{table}

\subsection{VLA Policy improvement via RL in WMs}
\label{sec:exp_rl}

Results in Table~\ref{tab:main_rl} translate the WM fidelity analysis into policy performance.
We compare our WM-based RL methods against two categories of baselines: 
\textbf{No-RL baselines}: the LIBERO-90 zero-shot policy and the 1-shot SFT policy.
\textbf{Online RL in the real simulator}: 
using ground-truth success rewards, 
we run GRPO with two real-rollout budgets---Online RL (Short) uses $\sim$500 target-suite episodes (matching the ID-FT WM data budget), 
and Online RL (Long) uses $\sim$2{,}500 episodes (matching WoVR's data budget).
For WM-based RL, we evaluate under the same four WM conditions as in Table~\ref{tab:wm_quality}, 
all using the zero-shot Qwen3-VL reward for fair comparison. 
Online RL and WM-based RL are both initialized from the 1-shot SFT policy. 
More details are in Appendix~\ref{sec:appendix_impl}.

Across all suites, every WM-based RL variant yields a clear improvement over the 1-shot SFT baseline, 
confirming that task-agnostic world modeling combined with a VLM reward is sufficient to drive meaningful RL gains on unseen tasks.
Using a zero-shot WM and a VLM reward, 
our method already outperforms Online RL (Short), which requires \(\sim\)500 real-rollout with ground-truth rewards (52.3\% v.s 47.9\%).
Co-Train WM RL approaches the performance of Online RL (Long) that consumes $\sim$2{,}500 real episodes, 
while incurring zero additional data collection beyond the 10 demonstrations already used for SFT (57.1\% v.s 58.8\%).
With only 500 in-domain trajectories for WM fine-tuning, 
RL in ID-FT WM surpasses Online RL (Long) using $5\times$ real rollouts and gt rewards (66.0\% v.s 58.8\%).
These results validate that imagination-only RL can be a competitive substitute for costly online interaction 
even under strict target-suite data constraints.
Remarkably, ID-FT WM RL also outperforms RL in WoVR's WM, which is learned from scratch with massive target-suite data.
This highlights extreme data efficiency, 
proving that the solid foundation established by the pre-trained WM needs only minimal in-domain data to significantly benefit RL.

Furthermore, the improvement ordering of our WM variants (Zero-shot \(\rightarrow\) Co-Train \(\rightarrow\) ID-FT) 
consistently predicts the ordering of downstream RL gains, 
proving that WM fidelity is a ceiling for imagination-based policy optimization.
The Object suite reflects this most sharply: 
the Zero-Shot WM has the worst simulation quality (FVD=233) and correspondingly shows almost no RL gain (+0.8\%), 
but as Co-Train and ID-FT improve prediction fidelity, 
the policy success rate increases dramatically (+13.8\% and +33.4\%).
On Long suite, without any target rollouts for the WM or reward, 
RL in Zero-Shot WM even significantly outperforms long-horizon Online RL (+15.6\% v.s +2.2\%), 
showcasing the power of the pretrained WM's transferable physical dynamics on hard, long-horizon tasks.
Notably, on Goal suite, the zero-shot VLM reward quality degrades (as shown in Appendix~\ref{sec:appendix_reward_eval}) 
and directly limits the policy improvement, but still improves upon the 1-shot SFT baseline.

\subsection{Ablation studies}
\label{sec:exp_ablation}

\begin{table}[t]
  \caption{\textbf{Component ablation.}
  We ablate the reward model and DNV under two WM conditions.
  ``1-shot RM'' is a VideoMAE-based classifier fine-tuned on the 10 1-shot SFT demonstrations per suite. 
  Robometer serves as an oracle reference, trained on LIBERO data that includes the target suites.
  DNV status is indicated per row.
  Bold rows correspond to our full method in Table~\ref{tab:main_rl}.}
  \label{tab:ablation}
  \centering
  \small
  \begin{tabular}{ll cccc c}
    \toprule
    WM & Configuration & Spatial & Object & Goal & Long & Avg. \\
    \midrule
    \multirow{4}{*}{Zero-Shot}
    & \textbf{Qwen3-VL, w/ DNV} & 65.8\,{\scriptsize(+11.2)} & 47.2\,{\scriptsize(+0.8)} & 60.2\,{\scriptsize(+8.0)} & \textbf{35.8}\,{\scriptsize\textbf{(+15.6)}} & \textbf{52.3}\,{\scriptsize\textbf{(+8.9)}} \\
    & Qwen3-VL, w/o DNV & 64.0\,{\scriptsize(+9.4)} & 46.0\,{\scriptsize(-0.4)} & 56.8\,{\scriptsize(+4.6)} & 30.8\,{\scriptsize(+10.6)} & 49.4\,{\scriptsize(+6.0)} \\
    & 1-shot RM, w/o DNV      & 50.6\,{\scriptsize(-4.0)} & 23.2\,{\scriptsize(-23.2)} & 37.2\,{\scriptsize(-15.0)} & 15.6\,{\scriptsize(-4.6)} & 31.7\,{\scriptsize(-11.7)} \\
    & Robometer, w/o DNV    & \textbf{68.4}\,{\scriptsize\textbf{(+13.8)}} & \textbf{50.0}\,{\scriptsize\textbf{(+3.6)}} & \textbf{61.8}\,{\scriptsize\textbf{(+9.6)}} & 22.0\,{\scriptsize(+1.8)} & 50.6\,{\scriptsize(+7.2)} \\
    \midrule
    \multirow{4}{*}{Co-Train}
    & \textbf{Qwen3-VL, w/ DNV} & \textbf{73.2}\,{\scriptsize\textbf{(+18.6)}} & 60.2\,{\scriptsize(+13.8)} & \textbf{58.4}\,{\scriptsize\textbf{(+6.2)}} & \textbf{36.6}\,{\scriptsize\textbf{(+16.4)}} & \textbf{57.1}\,{\scriptsize\textbf{(+13.7)}} \\
    & Qwen3-VL, w/o DNV & 72.4\,{\scriptsize(+17.8)} & 56.4\,{\scriptsize(+10.0)} & 58.2\,{\scriptsize(+6.0)} & 30.8\,{\scriptsize(+10.6)} & 54.5\,{\scriptsize(+11.1)} \\
    & 1-shot RM, w/o DNV      & 49.4\,{\scriptsize(-5.2)} & 45.8\,{\scriptsize(-0.6)} & 44.0\,{\scriptsize(-8.2)} & 20.8\,{\scriptsize(+0.6)} & 40.0\,{\scriptsize(-3.4)} \\
    & Robometer, w/o DNV    & 73.0\,{\scriptsize(+18.4)} & \textbf{64.8}\,{\scriptsize\textbf{(+18.4)}}  & 53.2\,{\scriptsize(+1.0)} & 21.8\,{\scriptsize(+1.6)} &  53.2\,{\scriptsize(+9.8)} \\
    \bottomrule
  \end{tabular}
\end{table}

We ablate the reward model usage and the DNV mechanism under Zero-Shot and Co-Train WM.
A central question of our task-agnostic paradigm is that,
when only minimal target-task data (one expert demo per task) is available,
whether a zero-shot VLM reward can replace the finetuned reward model as a success classifier.
Since DNV itself relies on the reward signal for its second-pass judgment, 
we first compare reward models \emph{without} DNV to isolate the reward signal's intrinsic quality.
We evaluate three reward variants:
\textbf{(i)}~\emph{Qwen3-VL} (zero-shot): our default, requiring no target-task data. 
\textbf{(ii)}~\emph{1-shot RM}: Following ~\cite{zhu2025wmpo}, 
we finetune a binary classifier from VideoMAE~\cite{tong2022videomae} for each suite, 
using the 10 expert demonstrations in 1-shot SFT.
We construct a small supervised dataset by sampling terminal segments of each demonstration as positive examples 
and early or cross-task segments as negatives, learning a multi-task reward model per suite from minimal data. 
\textbf{(iii)}~\emph{Robometer}~\cite{liang2026robometer} (oracle): 
a Qwen3-VL-4B-Instruct-based reward model finetuned on RBM-1M robotic dataset including LIBERO data, 
serving as an oracle upper bound with access to rich in-domain knowledge.

The results are in Table~\ref{tab:ablation}. 
The zero-shot Qwen3-VL reward significantly outperforms the 1-shot finetuned RM.
With only 10 positive examples, the learned classifier overfits to expert demonstration visuals 
and catastrophically misjudges the diverse, often imperfect outputs of the WM, 
actively degrading the policy below the SFT baseline on every suite.
By contrast, Qwen3-VL's broad visual commonsense generalizes robustly, 
performing comparably to the oracle Robometer.
When in-domain data is scarce, foundation VLM commonsense is far more reliable than a few-shot learned classifier.

We then ablate the effect of DNV with the reward fixed to Qwen3-VL.
DNV yields consistent gains, particularly on Libero-Long.
Since stochasticity and prediction uncertainty compound over autoregressive generation, 
long-horizon rollouts are most prone to hallucinations, which are effectively filtered by DNV, stabilizing RL.
Notably, combining Qwen3-VL with DNV approximates or outperforms the oracle Robometer reward.
However, DNV offers limited help when the upstream signal is itself unreliable. 
On Object under zero-shot WM, the simulation quality is too poor for any rollout to be faithful;
on Goal the VLM reward quality degrades (Appendix ~\ref{sec:appendix_reward_eval}), resulting in modest performance gain under Co-Train WM. More results of DNV can be found in Appendix~\ref{sec:appendix_ghost_analysis}.

\begin{figure}[t]
    \centering
    \includegraphics[width=\textwidth]{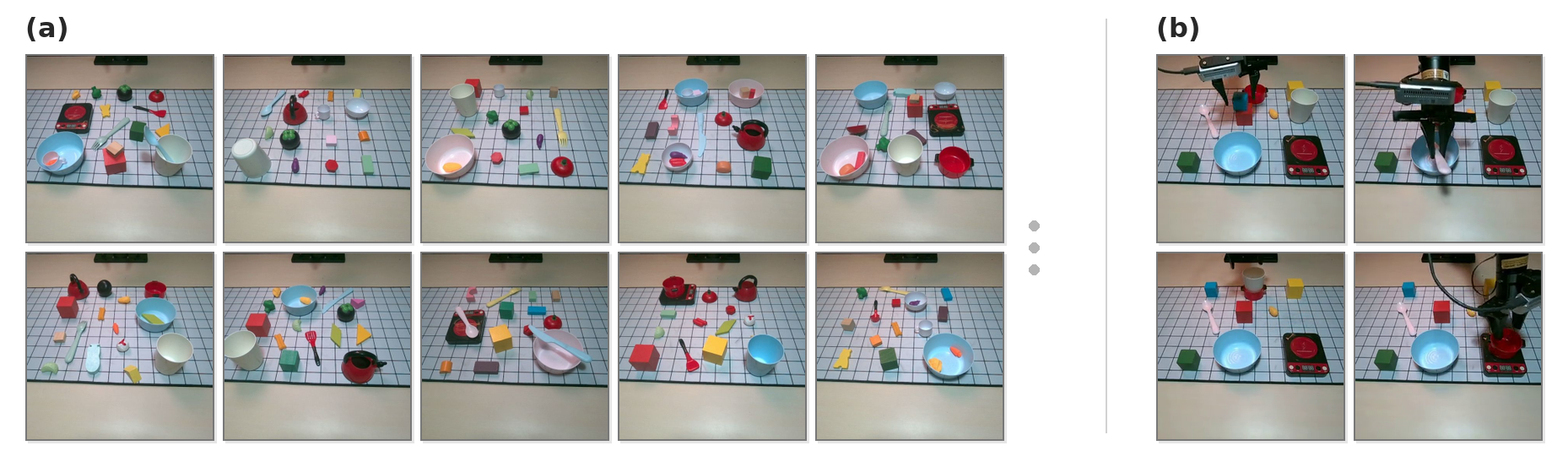}
    \caption{(a) Sample scenes from our collected play data spanning diverse object arrangements and tabletop layouts. (b) The four downstream evaluation tasks for VLA fine-tuning and RL.}
    \label{fig:piper}
\vskip -5pt
\end{figure}

\subsection{Real-World Experiments}
\label{sec:exp_real_world}

We deploy our pipeline and RAW-Dream on an AgileX Piper robotic arm with 7-DoF continuous control 
(6-DoF end-effector poses + 1-DoF continuous gripper within $[0, 0.1]$).
Our evaluation strictly adheres to the zero-target-data paradigm: 
the WM is never exposed to any downstream task data, 
and the VLA policy is improved from extremely-few-shot SFT base models entirely through imagination.

\textbf{Task-agnostic WM from play data.}
We first pre-train our WM on Open X-Embodiment datasets \cite{open_x_embodiment_rt_x_2023}, which is then fine-tuned on approximately 4 hours of uncurated, teleoperated play data, 
collected via a master-slave control system.
The tabletop workspace contains a rich variety of everyday objects 
arranged across different scene layouts with significantly randomized object positions and compositions, as is illustrated in Figure~\ref{fig:piper} (a).
During collection, the human operator freely performs whatever manipulation comes to mind
---grasping, placing, inserting, pushing, tumbling---without following any task definition or success criterion.
The data provides a generalizable physical prior, 
capturing a broad distribution of arm-object dynamics, contact patterns, and recovery behaviors,
while remaining agnostic to the downstream evaluation tasks.
We provide qualitative WM rollout results on unseen downstream scene layouts during WM training in Appendix~\ref{sec:appendix_realworld_rollout}.

\textbf{Evaluation protocols and Results. }
We evaluate on four contact-rich manipulation tasks in entirely unseen scene layouts: 
\emph{(1)} \textbf{Stack Block} (stack the blue block onto the red block), 
\emph{(2)} \textbf{Place Pot} (place the pot on the stove), 
\emph{(3)} \textbf{Put Spoon} (pick up the spoon and put it into the bowl), 
and \emph{(4)} \textbf{Place Cup} (place the cup into the pot), illustrated in Figure~\ref{fig:piper} (b). 
While similar manipulation semantics may appear in the play data, the specific scene layouts are entirely novel. 
The VLA policy is first anchored to each task via 3-shot SFT on teleoperated demonstrations, 
then improved via RL entirely within the play-data WM using zero-shot Qwen3-VL rewards, 
where the zero-shot WM is not finetined any more.
Each method is evaluated over 30 real-world trials per task.

\textbf{Results. }
As shown in Table~\ref{tab:real_world}, 
starting from a 3-shot SFT baseline at only 50.0\% average success rate, 
RAW-Dream yields a striking \textbf{+21.7\%} absolute gain (50.0\% $\rightarrow$ 71.7\%), 
with improvements on every task.
The largest gains appear on Put Spoon (+30.0\%) and Place Cup (+30.0\%), 
where compounding execution errors most severely limit the few-shot SFT policy.
Notably, all gains are achieved using only 3 teleoperated demonstrations per task for SFT 
and a fully frozen, play-data-only WM with zero-shot VLM rewards. No rollouts from the downstream scene are collected for either the WM or the reward.
These results confirm that broad physical priors captured by a downstream-agnostic WM are sufficient to drive reliable real-world policy improvement entirely through imagination. We provide video demonstrations of our real-world experiments on \href{https://sites.google.com/view/raw-dream-}{our website}.

\begin{table}[ht]
\centering
\caption{Real-world success rates (\%, 30 trials per task) on the AgileX Piper robot arm.}
\label{tab:real_world}
\small
\setlength{\tabcolsep}{4pt}
\renewcommand{\arraystretch}{1.15}
\begin{tabular}{l c c c c c}
\toprule
\textbf{Method} & \textbf{Stack block} & \textbf{Place pot} & \textbf{Put spoon} & \textbf{Place cup} & \textbf{Average} \\
\midrule
3-shot SFT
& 40.0 (12/30)
& 73.3 (22/30)
& 36.7 (11/30)
& 50.0 (15/30)
& 50.0 \\
RL from 3-shot SFT
& \textbf{50.0} (15/30)
& \textbf{90.0} (27/30)
& \textbf{66.7} (20/30)
& \textbf{80.0} (24/30)
& \textbf{71.7} \\
\bottomrule
\end{tabular}
\end{table}

\section{Conclusion}
\label{sec:conclusion}

We propose RAW-Dream, a paradigm enabling WM-based RL post-training of VLAs on unseen tasks. 
By combining a task-agnostic WM with a zero-shot VLM reward, RAW-Dream eliminates the need for massive in-domain data and target-task exposure. 
A dual-noise verification mechanism further stabilizes RL by filtering WM hallucinations. 
Experiments demonstrate that our method consistently outperforms SFT baselines
and data-heavy paradigms like online RL or RL in WMs learned from scratch,
unlocking a highly data-efficient and scalable pathway for WM-based VLA RL post-training.
Promising future directions include improving zero-shot WM fidelity by scaling model capacity and data diversity, 
enhancing zero-shot VLM reward quality through lightweight calibration, 
and extending our paradigm to flow-based VLA architectures.



\bibliographystyle{neurips}
\bibliography{neurips_2026}

@article{bai2025qwen3,
  title={Qwen3-vl technical report},
  author={Bai, Shuai and Cai, Yuxuan and Chen, Ruizhe and Chen, Keqin and Chen, Xionghui and Cheng, Zesen and Deng, Lianghao and Ding, Wei and Gao, Chang and Ge, Chunjiang and others},
  journal={arXiv preprint arXiv:2511.21631},
  year={2025}
}

@article{kim2025fine,
  title={Fine-tuning vision-language-action models: Optimizing speed and success},
  author={Kim, Moo Jin and Finn, Chelsea and Liang, Percy},
  journal={arXiv preprint arXiv:2502.19645},
  year={2025}
}

@article{zhu2025wmpo,
  title={Wmpo: World model-based policy optimization for vision-language-action models},
  author={Zhu, Fangqi and Yan, Zhengyang and Hong, Zicong and Shou, Quanxin and Ma, Xiao and Guo, Song},
  journal={arXiv preprint arXiv:2511.09515},
  year={2025}
}

@article{jiang2026wovr,
  title={Wovr: World models as reliable simulators for post-training vla policies with rl},
  author={Jiang, Zhennan and Zhou, Shangqing and Jiang, Yutong and Huang, Zefang and Wei, Mingjie and Chen, Yuhui and Zhou, Tianxing and Guo, Zhen and Lin, Hao and Zhang, Quanlu and others},
  journal={arXiv preprint arXiv:2602.13977},
  year={2026}
}

@article{liu2026world,
  title={World-VLA-Loop: Closed-Loop Learning of Video World Model and VLA Policy},
  author={Liu, Xiaokang and Bai, Zechen and Ci, Hai and Ma, Kevin Yuchen and Shou, Mike Zheng},
  journal={arXiv preprint arXiv:2602.06508},
  year={2026}
}

@article{sharma2026world,
  title={World-Gymnast: Training Robots with Reinforcement Learning in a World Model},
  author={Sharma, Ansh Kumar and Sun, Yixiang and Lu, Ninghao and Zhang, Yunzhe and Liu, Jiarao and Yang, Sherry},
  journal={arXiv preprint arXiv:2602.02454},
  year={2026}
}

@article{guo2026vlaw,
  title={Vlaw: Iterative co-improvement of vision-language-action policy and world model},
  author={Guo, Yanjiang and Lee, Tony and Shi, Lucy Xiaoyang and Chen, Jianyu and Liang, Percy and Finn, Chelsea},
  journal={arXiv preprint arXiv:2602.12063},
  year={2026}
}

@article{zhang2025reinforcing,
  title={Reinforcing action policies by prophesying},
  author={Zhang, Jiahui and Huang, Ze and Gu, Chun and Ma, Zipei and Zhang, Li},
  journal={arXiv preprint arXiv:2511.20633},
  year={2025}
}

@article{yin2026playworld,
  title={PlayWorld: Learning Robot World Models from Autonomous Play},
  author={Yin, Tenny and Mei, Zhiting and Zheng, Zhonghe and Yamane, Miyu and Wang, David and Sceats, Jade and Bateman, Samuel M and Zha, Lihan and Badithela, Apurva and Shorinwa, Ola and others},
  journal={arXiv preprint arXiv:2603.09030},
  year={2026}
}

@article{zhang2026towards,
  title={Towards Practical World Model-based Reinforcement Learning for Vision-Language-Action Models},
  author={Zhang, Zhilong and Ren, Haoxiang and Sun, Yihao and Sheng, Yifei and Wang, Haonan and Lin, Haoxin and Wu, Zhichao and Bacon, Pierre-Luc and Yu, Yang},
  journal={arXiv preprint arXiv:2603.20607},
  year={2026}
}

@article{yang2026rise,
  title={Rise: Self-improving robot policy with compositional world model},
  author={Yang, Jiazhi and Lin, Kunyang and Li, Jinwei and Zhang, Wencong and Lin, Tianwei and Wu, Longyan and Su, Zhizhong and Zhao, Hao and Zhang, Ya-Qin and Chen, Li and others},
  journal={arXiv preprint arXiv:2602.11075},
  year={2026}
}

@article{li2025vla,
  title={Vla-rft: Vision-language-action reinforcement fine-tuning with verified rewards in world simulators},
  author={Li, Hengtao and Ding, Pengxiang and Suo, Runze and Wang, Yihao and Ge, Zirui and Zang, Dongyuan and Yu, Kexian and Sun, Mingyang and Zhang, Hongyin and Wang, Donglin and others},
  journal={arXiv preprint arXiv:2510.00406},
  year={2025}
}

@article{hung2025nora,
  title={Nora-1.5: A vision-language-action model trained using world model-and action-based preference rewards},
  author={Hung, Chia-Yu and Majumder, Navonil and Deng, Haoyuan and Renhang, Liu and Ang, Yankang and Zadeh, Amir and Li, Chuan and Herremans, Dorien and Wang, Ziwei and Poria, Soujanya},
  journal={arXiv preprint arXiv:2511.14659},
  year={2025}
}

@article{xiao2025world,
  title={World-env: Leveraging world model as a virtual environment for vla post-training},
  author={Xiao, Junjin and Yang, Yandan and Chang, Xinyuan and Chen, Ronghan and Xiong, Feng and Xu, Mu and Zheng, Wei-Shi and Zhang, Qing},
  journal={arXiv preprint arXiv:2509.24948},
  year={2025}
}

@article{liang2026robometer,
  title={Robometer: Scaling general-purpose robotic reward models via trajectory comparisons},
  author={Liang, Anthony and Korkmaz, Yigit and Zhang, Jiahui and Hwang, Minyoung and Anwar, Abrar and Kaushik, Sidhant and Shah, Aditya and Huang, Alex S and Zettlemoyer, Luke and Fox, Dieter and others},
  journal={arXiv preprint arXiv:2603.02115},
  year={2026}
}

@article{lu2025vla,
  title={Vla-rl: Towards masterful and general robotic manipulation with scalable reinforcement learning},
  author={Lu, Guanxing and Guo, Wenkai and Zhang, Chubin and Zhou, Yuheng and Jiang, Haonan and Gao, Zifeng and Tang, Yansong and Wang, Ziwei},
  journal={arXiv preprint arXiv:2505.18719},
  year={2025}
}

@article{chen202500_0texttt0rl000,
  title   = {$\pi_\texttt{RL}$: Online RL Fine-tuning for Flow-based Vision-Language-Action Models},
  author  = {Kang Chen and Zhihao Liu and Tonghe Zhang and Zhen Guo and Si Xu and Hao Lin and Hongzhi Zang and Quanlu Zhang and Zhaofei Yu and Guoliang Fan and Tiejun Huang and Yu Wang and Chao Yu},
  year    = {2025},
  journal = {arXiv preprint arXiv: 2510.25889}
}

@article{intelligence2025000000_0006000,
  title   = {$\pi^{*}_{0.6}$: a VLA That Learns From Experience},
  author  = {Physical Intelligence and Ali Amin and Raichelle Aniceto and Ashwin Balakrishna and Kevin Black and Ken Conley and Grace Connors and James Darpinian and Karan Dhabalia and Jared DiCarlo and Danny Driess and Michael Equi and Adnan Esmail and Yunhao Fang and Chelsea Finn and Catherine Glossop and Thomas Godden and Ivan Goryachev and Lachy Groom and Hunter Hancock and Karol Hausman and Gashon Hussein and Brian Ichter and Szymon Jakubczak and Rowan Jen and Tim Jones and Ben Katz and Liyiming Ke and Chandra Kuchi and Marinda Lamb and Devin LeBlanc and Sergey Levine and Adrian Li-Bell and Yao Lu and Vishnu Mano and Mohith Mothukuri and Suraj Nair and Karl Pertsch and Allen Z. Ren and Charvi Sharma and Lucy Xiaoyang Shi and Laura Smith and Jost Tobias Springenberg and Kyle Stachowicz and Will Stoeckle and Alex Swerdlow and James Tanner and Marcel Torne and Quan Vuong and Anna Walling and Haohuan Wang and Blake Williams and Sukwon Yoo and Lili Yu and Ury Zhilinsky and Zhiyuan Zhou},
  year    = {2025},
  journal = {arXiv preprint arXiv: 2511.14759}
}

@article{xu2026rl,
  title={RL Token: Bootstrapping Online RL with Vision-Language-Action Models},
  author={Xu, Charles and Springenberg, Jost Tobias and Equi, Michael and Amin, Ali and Esmail, Adnan and Levine, Sergey and Ke, Liyiming},
  journal={arXiv preprint arXiv:2604.23073},
  year={2026}
}

@article{li2025simplevla,
  title={Simplevla-rl: Scaling vla training via reinforcement learning},
  author={Li, Haozhan and Zuo, Yuxin and Yu, Jiale and Zhang, Yuhao and Yang, Zhaohui and Zhang, Kaiyan and Zhu, Xuekai and Zhang, Yuchen and Chen, Tianxing and Cui, Ganqu and others},
  journal={arXiv preprint arXiv:2509.09674},
  year={2025}
}

@article{shao2024deepseekmath,
  title={Deepseekmath: Pushing the limits of mathematical reasoning in open language models},
  author={Shao, Zhihong and Wang, Peiyi and Zhu, Qihao and Xu, Runxin and Song, Junxiao and Bi, Xiao and Zhang, Haowei and Zhang, Mingchuan and Li, YK and Wu, Yang and others},
  journal={arXiv preprint arXiv:2402.03300},
  year={2024}
}

@article{team2025evaluating,
  title   = {Evaluating Gemini Robotics Policies in a Veo World Simulator},
  author  = {Gemini Robotics Team and Coline Devin and Yilun Du and Debidatta Dwibedi and Ruiqi Gao and Abhishek Jindal and Thomas Kipf and Sean Kirmani and Fangchen Liu and Anirudha Majumdar and Andrew Marmon and Carolina Parada and Yulia Rubanova and Dhruv Shah and Vikas Sindhwani and Jie Tan and Fei Xia and Ted Xiao and Sherry Yang and Wenhao Yu and Allan Zhou},
  year    = {2025},
  journal = {arXiv preprint arXiv: 2512.10675}
}

@article{quevedo2025worldgym0,
  title   = {WorldGym: World Model as An Environment for Policy Evaluation},
  author  = {Julian Quevedo and Ansh Kumar Sharma and Yixiang Sun and Varad Suryavanshi and Percy Liang and Sherry Yang},
  year    = {2025},
  journal = {arXiv preprint arXiv: 2506.00613}
}

@article{tseng2025scalable,
  title   = {Scalable Policy Evaluation with Video World Models},
  author  = {Wei-Cheng Tseng and Jinwei Gu and Qinsheng Zhang and Hanzi Mao and Ming-Yu Liu and Florian Shkurti and Lin Yen-Chen},
  year    = {2025},
  journal = {arXiv preprint arXiv: 2511.11520}
}

@article{wan2025wan,
  title={Wan: Open and advanced large-scale video generative models},
  author={Wan, Team and Wang, Ang and Ai, Baole and Wen, Bin and Mao, Chaojie and Xie, Chen-Wei and Chen, Di and Yu, Feiwu and Zhao, Haiming and Yang, Jianxiao and others},
  journal={arXiv preprint arXiv:2503.20314},
  year={2025}
}

@article{chandra2025diwa,
  title={Diwa: Diffusion policy adaptation with world models},
  author={Chandra, Akshay L and Nematollahi, Iman and Huang, Chenguang and Welschehold, Tim and Burgard, Wolfram and Valada, Abhinav},
  journal={arXiv preprint arXiv:2508.03645},
  year={2025}
}

@article{mazzaglia2024genrl,
  title={Genrl: Multimodal-foundation world models for generalization in embodied agents},
  author={Mazzaglia, Pietro and Verbelen, Tim and Dhoedt, Bart and Courville, Aaron and Rajeswar, Sai},
  journal={Advances in neural information processing systems},
  volume={37},
  pages={27529--27555},
  year={2024}
}

@article{wang2025founder,
  title={Founder: Grounding foundation models in world models for open-ended embodied decision making},
  author={Wang, Yucen and Yu, Rui and Wan, Shenghua and Gan, Le and Zhan, De-Chuan},
  journal={arXiv preprint arXiv:2507.12496},
  year={2025}
}

@inproceedings{sekar2020planning,
  title={Planning to explore via self-supervised world models},
  author={Sekar, Ramanan and Rybkin, Oleh and Daniilidis, Kostas and Abbeel, Pieter and Hafner, Danijar and Pathak, Deepak},
  booktitle={International conference on machine learning},
  pages={8583--8592},
  year={2020},
  organization={PMLR}
}

@article{lu2022challenges,
  title={Challenges and opportunities in offline reinforcement learning from visual observations},
  author={Lu, Cong and Ball, Philip J and Rudner, Tim GJ and Parker-Holder, Jack and Osborne, Michael A and Teh, Yee Whye},
  journal={arXiv preprint arXiv:2206.04779},
  year={2022}
}

@article{liu2023libero,
  title={Libero: Benchmarking knowledge transfer for lifelong robot learning},
  author={Liu, Bo and Zhu, Yifeng and Gao, Chongkai and Feng, Yihao and Liu, Qiang and Zhu, Yuke and Stone, Peter},
  journal={Advances in Neural Information Processing Systems},
  volume={36},
  pages={44776--44791},
  year={2023}
}

@inproceedings{peebles2023scalable,
  title={Scalable diffusion models with transformers},
  author={Peebles, William and Xie, Saining},
  booktitle={Proceedings of the IEEE/CVF international conference on computer vision},
  pages={4195--4205},
  year={2023}
}

@inproceedings{he2026pre,
  title={Pre-trained video generative models as world simulators},
  author={He, Haoran and Zhang, Yang and Lin, Liang and Xu, Zhongwen and Pan, Ling},
  booktitle={Proceedings of the AAAI Conference on Artificial Intelligence},
  volume={40},
  pages={4645--4653},
  year={2026}
}

@inproceedings{zhu2025irasim,
  title={Irasim: A fine-grained world model for robot manipulation},
  author={Zhu, Fangqi and Wu, Hongtao and Guo, Song and Liu, Yuxiao and Cheang, Chilam and Kong, Tao},
  booktitle={Proceedings of the IEEE/CVF International Conference on Computer Vision},
  pages={9834--9844},
  year={2025}
}

@article{wang2025co,
  title={Co-Evolving Latent Action World Models},
  author={Wang, Yucen and Zhang, Fengming and Zhan, De-Chuan and Zhao, Li and Wang, Kaixin and Bian, Jiang},
  journal={arXiv preprint arXiv:2510.26433},
  year={2025}
}

@article{ali2025world,
  title={World simulation with video foundation models for physical ai},
  author={Ali, Arslan and Bai, Junjie and Bala, Maciej and Balaji, Yogesh and Blakeman, Aaron and Cai, Tiffany and Cao, Jiaxin and Cao, Tianshi and Cha, Elizabeth and Chao, Yu-Wei and others},
  journal={arXiv preprint arXiv:2511.00062},
  year={2025}
}

@article{yu2020mopo,
  title={Mopo: Model-based offline policy optimization},
  author={Yu, Tianhe and Thomas, Garrett and Yu, Lantao and Ermon, Stefano and Zou, James Y and Levine, Sergey and Finn, Chelsea and Ma, Tengyu},
  journal={Advances in neural information processing systems},
  volume={33},
  pages={14129--14142},
  year={2020}
}

@article{kidambi2020morel,
  title={Morel: Model-based offline reinforcement learning},
  author={Kidambi, Rahul and Rajeswaran, Aravind and Netrapalli, Praneeth and Joachims, Thorsten},
  journal={Advances in neural information processing systems},
  volume={33},
  pages={21810--21823},
  year={2020}
}

@article{black2024pi0,
  title={$\pi_0$: A Vision-Language-Action Flow Model for General Robot Control},
  author={Black, Kevin and Brown, Noah and Driess, Danny and Esmail, Adnan and Equi, Michael and Finn, Chelsea and Fusai, Niccolo and Groom, Lachy and Hausman, Karol and Ichter, Brian and others},
  journal={arXiv preprint arXiv:2410.24164},
  year={2024}
}

@article{intelligence2025pi,
  title={$\pi_{0.5}$: a Vision-Language-Action Model with Open-World Generalization},
  author={Intelligence, Physical and Black, Kevin and Brown, Noah and Darpinian, James and Dhabalia, Karan and Driess, Danny and Esmail, Adnan and Equi, Michael and Finn, Chelsea and Fusai, Niccolo and others},
  journal={arXiv preprint arXiv:2504.16054},
  year={2025}
}

@article{kim2024openvla,
  title={Openvla: An open-source vision-language-action model},
  author={Kim, Moo Jin and Pertsch, Karl and Karamcheti, Siddharth and Xiao, Ted and Balakrishna, Ashwin and Nair, Suraj and Rafailov, Rafael and Foster, Ethan and Lam, Grace and Sanketi, Pannag and others},
  journal={arXiv preprint arXiv:2406.09246},
  year={2024}
}

@article{chen2025villa,
  title={Villa-x: enhancing latent action modeling in vision-language-action models},
  author={Chen, Xiaoyu and Wei, Hangxing and Zhang, Pushi and Zhang, Chuheng and Wang, Kaixin and Guo, Yanjiang and Yang, Rushuai and Wang, Yucen and Xiao, Xinquan and Zhao, Li and others},
  journal={arXiv preprint arXiv:2507.23682},
  year={2025}
}

@article{tong2022videomae,
  title={Videomae: Masked autoencoders are data-efficient learners for self-supervised video pre-training},
  author={Tong, Zhan and Song, Yibing and Wang, Jue and Wang, Limin},
  journal={Advances in neural information processing systems},
  volume={35},
  pages={10078--10093},
  year={2022}
}

@misc{open_x_embodiment_rt_x_2023,
title={Open {X-E}mbodiment: Robotic Learning Datasets and {RT-X} Models},
author = {Open X-Embodiment Collaboration and Abby O'Neill and Abdul Rehman and Abhiram Maddukuri and Abhishek Gupta and Abhishek Padalkar and Abraham Lee and Acorn Pooley and Agrim Gupta and Ajay Mandlekar and Ajinkya Jain and Albert Tung and Alex Bewley and Alex Herzog and Alex Irpan and Alexander Khazatsky and Anant Rai and Anchit Gupta and Andrew Wang and Andrey Kolobov and Anikait Singh and Animesh Garg and Aniruddha Kembhavi and Annie Xie and Anthony Brohan and Antonin Raffin and Archit Sharma and Arefeh Yavary and Arhan Jain and Ashwin Balakrishna and Ayzaan Wahid and Ben Burgess-Limerick and Beomjoon Kim and Bernhard Schölkopf and Blake Wulfe and Brian Ichter and Cewu Lu and Charles Xu and Charlotte Le and Chelsea Finn and Chen Wang and Chenfeng Xu and Cheng Chi and Chenguang Huang and Christine Chan and Christopher Agia and Chuer Pan and Chuyuan Fu and Coline Devin and Danfei Xu and Daniel Morton and Danny Driess and Daphne Chen and Deepak Pathak and Dhruv Shah and Dieter Büchler and Dinesh Jayaraman and Dmitry Kalashnikov and Dorsa Sadigh and Edward Johns and Ethan Foster and Fangchen Liu and Federico Ceola and Fei Xia and Feiyu Zhao and Felipe Vieira Frujeri and Freek Stulp and Gaoyue Zhou and Gaurav S. Sukhatme and Gautam Salhotra and Ge Yan and Gilbert Feng and Giulio Schiavi and Glen Berseth and Gregory Kahn and Guangwen Yang and Guanzhi Wang and Hao Su and Hao-Shu Fang and Haochen Shi and Henghui Bao and Heni Ben Amor and Henrik I Christensen and Hiroki Furuta and Homer Walke and Hongjie Fang and Huy Ha and Igor Mordatch and Ilija Radosavovic and Isabel Leal and Jacky Liang and Jad Abou-Chakra and Jaehyung Kim and Jaimyn Drake and Jan Peters and Jan Schneider and Jasmine Hsu and Jeannette Bohg and Jeffrey Bingham and Jeffrey Wu and Jensen Gao and Jiaheng Hu and Jiajun Wu and Jialin Wu and Jiankai Sun and Jianlan Luo and Jiayuan Gu and Jie Tan and Jihoon Oh and Jimmy Wu and Jingpei Lu and Jingyun Yang and Jitendra Malik and João Silvério and Joey Hejna and Jonathan Booher and Jonathan Tompson and Jonathan Yang and Jordi Salvador and Joseph J. Lim and Junhyek Han and Kaiyuan Wang and Kanishka Rao and Karl Pertsch and Karol Hausman and Keegan Go and Keerthana Gopalakrishnan and Ken Goldberg and Kendra Byrne and Kenneth Oslund and Kento Kawaharazuka and Kevin Black and Kevin Lin and Kevin Zhang and Kiana Ehsani and Kiran Lekkala and Kirsty Ellis and Krishan Rana and Krishnan Srinivasan and Kuan Fang and Kunal Pratap Singh and Kuo-Hao Zeng and Kyle Hatch and Kyle Hsu and Laurent Itti and Lawrence Yunliang Chen and Lerrel Pinto and Li Fei-Fei and Liam Tan and Linxi "Jim" Fan and Lionel Ott and Lisa Lee and Luca Weihs and Magnum Chen and Marion Lepert and Marius Memmel and Masayoshi Tomizuka and Masha Itkina and Mateo Guaman Castro and Max Spero and Maximilian Du and Michael Ahn and Michael C. Yip and Mingtong Zhang and Mingyu Ding and Minho Heo and Mohan Kumar Srirama and Mohit Sharma and Moo Jin Kim and Naoaki Kanazawa and Nicklas Hansen and Nicolas Heess and Nikhil J Joshi and Niko Suenderhauf and Ning Liu and Norman Di Palo and Nur Muhammad Mahi Shafiullah and Oier Mees and Oliver Kroemer and Osbert Bastani and Pannag R Sanketi and Patrick "Tree" Miller and Patrick Yin and Paul Wohlhart and Peng Xu and Peter David Fagan and Peter Mitrano and Pierre Sermanet and Pieter Abbeel and Priya Sundaresan and Qiuyu Chen and Quan Vuong and Rafael Rafailov and Ran Tian and Ria Doshi and Roberto Mart{'i}n-Mart{'i}n and Rohan Baijal and Rosario Scalise and Rose Hendrix and Roy Lin and Runjia Qian and Ruohan Zhang and Russell Mendonca and Rutav Shah and Ryan Hoque and Ryan Julian and Samuel Bustamante and Sean Kirmani and Sergey Levine and Shan Lin and Sherry Moore and Shikhar Bahl and Shivin Dass and Shubham Sonawani and Shuran Song and Sichun Xu and Siddhant Haldar and Siddharth Karamcheti and Simeon Adebola and Simon Guist and Soroush Nasiriany and Stefan Schaal and Stefan Welker and Stephen Tian and Subramanian Ramamoorthy and Sudeep Dasari and Suneel Belkhale and Sungjae Park and Suraj Nair and Suvir Mirchandani and Takayuki Osa and Tanmay Gupta and Tatsuya Harada and Tatsuya Matsushima and Ted Xiao and Thomas Kollar and Tianhe Yu and Tianli Ding and Todor Davchev and Tony Z. Zhao and Travis Armstrong and Trevor Darrell and Trinity Chung and Vidhi Jain and Vincent Vanhoucke and Wei Zhan and Wenxuan Zhou and Wolfram Burgard and Xi Chen and Xiangyu Chen and Xiaolong Wang and Xinghao Zhu and Xinyang Geng and Xiyuan Liu and Xu Liangwei and Xuanlin Li and Yansong Pang and Yao Lu and Yecheng Jason Ma and Yejin Kim and Yevgen Chebotar and Yifan Zhou and Yifeng Zhu and Yilin Wu and Ying Xu and Yixuan Wang and Yonatan Bisk and Yongqiang Dou and Yoonyoung Cho and Youngwoon Lee and Yuchen Cui and Yue Cao and Yueh-Hua Wu and Yujin Tang and Yuke Zhu and Yunchu Zhang and Yunfan Jiang and Yunshuang Li and Yunzhu Li and Yusuke Iwasawa and Yutaka Matsuo and Zehan Ma and Zhuo Xu and Zichen Jeff Cui and Zichen Zhang and Zipeng Fu and Zipeng Lin},
howpublished  = {\url{https://arxiv.org/abs/2310.08864}},
year = {2023},
}

@article{yu2025rlinf,
  title={Rlinf: Flexible and efficient large-scale reinforcement learning via macro-to-micro flow transformation},
  author={Yu, Chao and Wang, Yuanqing and Guo, Zhen and Lin, Hao and Xu, Si and Zang, Hongzhi and Zhang, Quanlu and Wu, Yongji and Zhu, Chunyang and Hu, Junhao and others},
  journal={arXiv preprint arXiv:2509.15965},
  year={2025}
}

@article{liu2022flow,
  title={Flow straight and fast: Learning to generate and transfer data with rectified flow},
  author={Liu, Xingchao and Gong, Chengyue and Liu, Qiang},
  journal={arXiv preprint arXiv:2209.03003},
  year={2022}
}


\clearpage

\appendix

\section{Implementation Details}
\label{sec:appendix_impl}

\subsection{Action-Conditioned World Model}
\label{sec:appendix_wm_details}

\paragraph{Architecture.}
We build on the WAN~2.1 T2V-1.3B DiT backbone with a paired VAE 
(latent dim $C{=}16$, stride $(4,8,8)$), 
yielding a $32{\times}32$ spatial latent grid from $256{\times}256$ pixel inputs 
and $4{\times}$ temporal compression (33 pixel frames $\to$ $\sim$8 latent frames per training window).
For action conditioning, 
each latent frame maps to a chunk of 4 raw action frames (matching the VAE temporal stride). 
Each chunk is projected by a from-scratch MLP into action tokens.
These enter every DiT block via AdaLN described in Section~\ref{sec:method_wm}.
Temporal self-attention is block-wise causal: 
spatial tokens within the same timestep attend bidirectionally, 
but across timesteps attention is strictly causal.

\paragraph{Training.}
We train on 33-frame windows.
Each window is partitioned into context frames and generation frames via a binary mask.
The context--generation split is randomized per sample.
Context frames are corrupted with per-position noise sampled from $\mathcal{U}[0, 300]$ out of 1000 total diffusion timesteps, 
training the model to predict from imperfect contexts.
We use AdamW with learning rate $7.5e{-5}$, linear warmup followed by cosine decay, 
batch size 12, and train for $\sim$100K steps.
The rectified flow objective supervises velocity prediction
with MSE loss computed only over valid generation positions.
Classifier-free guidance (CFG) dropout is disabled both in training and inference, which we found helpful for performance. Actions inputs are normalized on the WM training data via q01-q99 statistics.

\paragraph{Autoregressive inference.}
Each AR step generates $G{=}2$ latent states (8 pixel frames, matching the OpenVLA-OFT action chunk size) 
and conditions.
CFG is disabled on a window of 7 context frames: 
1 first-frame anchor + 6 recently generated latent states.
The AR iterations are performed entirely in latent space without intermediate VAE decode--encode round-trips.
Each denoising step uses the UniPC multi-step ODE solver with 5 function evaluations.
CFG is disabled.
Non-anchor context frames are injected at a fixed noise level $t_c{=}50$.
For the first-frame anchor, we apply \emph{progressive noise} starting from AR step $s_0{=}3$: 
the anchor noise level increases linearly as $t_1^{(s)} = \min(10 \cdot \max(s - 3, 0),\; 300)$.
The increasing noise gradually relaxes the anchor's influence.
This adjustment is applied \emph{only at inference} and does not alter the training.

\subsection{Dual-Noise Verification (DNV) in GRPO}
\label{sec:appendix_dnv}

To integrate DNV into GRPO, we handle unreliable rollouts as follows.
For each GRPO group of $G{=}8$ trajectories, 
any rollout whose VLM reward disagrees between the original and replayed generation 
has its reward replaced by the mean reward of the remaining trustworthy rollouts.
This forces the flagged trajectory's advantage to $\hat{A}_i = 0$, 
rendering it invisible to the policy gradient without changing the fixed group tensor size in RL optimization.
While this may shrink the standard deviation when computing advantages in the group of only reliable trajectories, 
the effect is merely a constant scaling factor on the gradients and does not harm optimization.
If more than two rollouts within a single group are flagged, 
the entire group is discarded to preserve optimization stability.

\subsection{VLA Policy and SFT Training}
\label{sec:appendix_vla_sft}

We adopt OpenVLA-OFT~\cite{kim2025fine} as the VLA backbone, 
which outputs an action chunk of 8 steps per inference call.
We do not use proprioceptive input or wrist camera images.
Unlike the original OpenVLA-OFT which uses deterministic L1 regression heads, 
we retain the probabilistic (categorical) action tokenization, 
as it naturally provides the stochastic policy required for GRPO sampling during RL.
For SFT on LIBERO-90, we use the full curated expert demonstrations and train for 150K steps.
For 1-shot SFT on each target suite, we fine-tune the LIBERO-90 SFT checkpoint 
on the 10 expert demonstrations (1 per task) for 50K steps.

\subsection{RL Training}
\label{sec:appendix_RL}

We use GRPO with group size $G{=}8$.
No KL reference penalty is applied ($\lambda_{\text{KL}}{=}0$).
Groups with all failure or all success rollouts are discarded during training. 
The VLA temperature when sampling actions during rollout is $1.6$. The learning rate is $5e{-6}$. All experiments are conducted using eight NVIDIA H200 GPUs. Training RL in our WM specifically takes about 2 days.

\textbf{WM-based RL.} Training steps per suite are: 20 for Object (early stopping since Object tends to produce all-success groups when we extend the training steps, which will be filtered, and GRPO will wait for a long time for a group that exhibits difference in rewards), and 50 for other three suites.

\textbf{Online RL.}
Both Online RL (Short) and Online RL (Long) use GRPO with the same OpenVLA-OFT architecture, 
initialized from the 1-shot SFT checkpoint.
Rollouts are collected in the real LIBERO simulator 
with ground-truth binary success rewards.
Online RL (Short) uses $\sim$500 episodes (512 episode, 1 online RL step) per suite comparable to the budget of ID-FT WM fine-tuning, and this setting is the same as WMPO; Online RL (Long) uses $\sim$2{,}500 episodes (2560 episodes, 10 online RL steps) comparable to the budget of WoVR WM training.

\textbf{WoVR.}
We compare against WoVR~\cite{jiang2026wovr}, which trains a WAN~2.2-TI2V-5B-based WM trained from scratch 
on 2,500 target-suite rollouts per suite without any broad pre-training, to validate the effect of the learned physical prior. 
We directly use their inference implementation in RL-Inf~\cite{yu2025rlinf} and their released WM checkpoints on the four LIBERO suites. To fairly evaluate the RL performance on different WM variants, we also use the VLM-based reward for RL training in WoVR WM. The VLM usage, inference parameters and the suite training steps are all the same with ours. 

\subsection{VLM Reward Deployment}
\label{sec:appendix_vlm_reward}

We deploy Qwen3-VL via vLLM as the zero-shot reward model.
Model selection per suite: Qwen3-VL-8B-Instruct for Spatial and Object; Qwen3-VL-32B-Instruct for Goal, Long and real-world tasks.
Inference configuration: temperature $0.7$, $n{=}5$ samples per request (majority vote), 
vote threshold $\geq 4/5$ for a positive reward.
Input videos are sampled at 4 FPS from the WM-generated rollout.

The prompt follows a simple task-agnostic template:
\begin{quote}
\texttt{
You are a highly rigorous robotics operations evaluation expert. Your task is to observe the provided video of a robot and objectively determine if it successfully completed the "\$\{task\_description\}" task. Please strictly output your evaluation in the following JSON format, without generating any extra text or markdown formatting:
  \{
      "Reasoning": "Briefly analyze whether the robot successfully completed the task.",
      "Final Answer": [Output ONLY one word: Success or Failure.]
  \}
}
\end{quote}
where the \$\{task\_description\} is the original description of the Libero tasks in simulation, and for real-world experiments we also directly use the task description mentioned in Section ~\ref{sec:exp_real_world}. We found that this minimal prompt outperforms more complex chain-of-thought variants 
that impose robot-specific evaluation criteria, 
as the latter tend to be overly strict and reduce recall.

\subsection{1-shot Fine-tuned Reward Model}
\label{sec:appendix_1shot_rm}

Following~\cite{zhu2025wmpo}, we finetune a VideoMAE~\cite{tong2022videomae} binary classifier per suite 
using the 10 expert demonstrations from 1-shot SFT, resulting in four multi-task binary reward models.
Positive examples are sampled from terminal segments of successful demonstrations; 
negative examples are constructed from early segments and cross-task segments. The length of each segment is 8. We add a learnable embedding for each task id in one suite to enable multi-task prediction ability. Furthermore, since the 1-shot finetuning quickly overfits to the scarce all-expert data, the originally-produced threshold ($\geq 0.9$, validated on these 10 training episodes since we have no access to more in-domain validation data) is proved to be so strict that nearly none of the  rollouts are predicted to be success. So we change to use a threshold of 0.5 in RL. During inference, the reward model scans the trajectory using a sliding window before finding a segment whose success probability exceeds the threshold (which is regarded as the terminal state of this success rollout, and the rest frames are discarded) . If none of segment is predited as success, the trajecory is predicted as a failure.

\subsection{Robometer}
\label{sec:appendix_robometer}

Robometer~\cite{liang2026robometer} is a VLM-based general-purpose robotic reward model trained on the RBM-1M dataset via a dual objective of frame-level progress prediction and trajectory-level preference comparison. It takes instruction and video as input and outputs dense progress and binary success labels in a zero-shot manner across tasks and embodiments.
In our experiments, we use it as an oracle reward upper bound. We deploy it as an HTTP service, using its predicted success probability for each subchunk to give binary rewards, with a threshold of 0.85. Similar to 1-shot fine-tuned reward model, if a subchunk is already judged as success before the end of the sequence, we discard the rest frames and regard this step as the terminal state for RL.

\section{Offline Evaluation of Reward Models}
\label{sec:appendix_reward_eval}

To validate the reliability of our task-agnostic reward, we conduct an offline evaluation of the reward models using a real-environment rollouts dataset on LIBERO (about 10K real trajectories for each of the four suites) with ground-truth binary success labels. We compare the Qwen3-VL alongside oracle Robometer-4B as a reference baseline. We do not compare with 1-shot in-domain finetuned reward model, since the results are too low and scanning along the whole trajectory is too slow for large-scale evaluation. The VLM inference parameters are all the same as in our RL training.

The evaluation reports the micro-averaged Precision (P), Recall (R), and micro-F1-score across all tasks in the four LIBERO suites. A robust reward model needs high precision to avoid false-positive gradients during RL, while maintaining enough recall to provide a meaningful learning signal.

As shown in Table~\ref{tab:offline_reward}, Qwen3-VL delivers strong and consistent F1-scores and Precision on Spatial, Object and Long, without requiring a single target-domain trajectory. While Robometer-4B offers near-oracle performance on the Object suite (likely due to the inclusion of similar prior data), its performance weirdly drops on Spatial and Long. Both models perform poor on Goal, which is the reason for the lowest gain in RL performance across the four suites, as shown in Table~\ref{tab:main_rl} and Table~\ref{tab:ablation}.

\begin{table}[h]
  \caption{Offline Reward Model Evaluation (Averaged precision, recall, and F1 across suites (\%)).}
  \label{tab:offline_reward}
  \centering
  \small
  \setlength{\tabcolsep}{4pt}
  \begin{tabular}{l ccc ccc}
    \toprule
    & \multicolumn{3}{c}{Qwen3-VL} & \multicolumn{3}{c}{Robometer} \\
    \cmidrule(lr){2-4} \cmidrule(lr){5-7} 
    Suite & P & R & F1 & P & R & F1  \\
    \midrule
    Spatial   & \textbf{85.8} & \textbf{80.0} & \textbf{82.8} & 73.5 & 20.1 & 31.5 \\
    Object    & 90.6 & 91.8 & 91.2 & \textbf{94.3} & \textbf{94.8} & \textbf{94.6}  \\
    Goal      & \textbf{61.1} & \textbf{24.5} & \textbf{35.0} & 27.5 & 8.9 & 13.4 \\
    Long & \textbf{62.0} & \textbf{85.8} & \textbf{72.0} & 53.8 & 73.9 & 62.2  \\
    \bottomrule
  \end{tabular}
\end{table}

\section{Additional Results and Analysis}

\subsection{Qualitative Analysis on First-Frame Ghosting}
\label{sec:appendix_ghost}
\begin{figure}[t]
    \centering
    \begin{subfigure}{\textwidth}
        \centering
        \includegraphics[width=\textwidth]{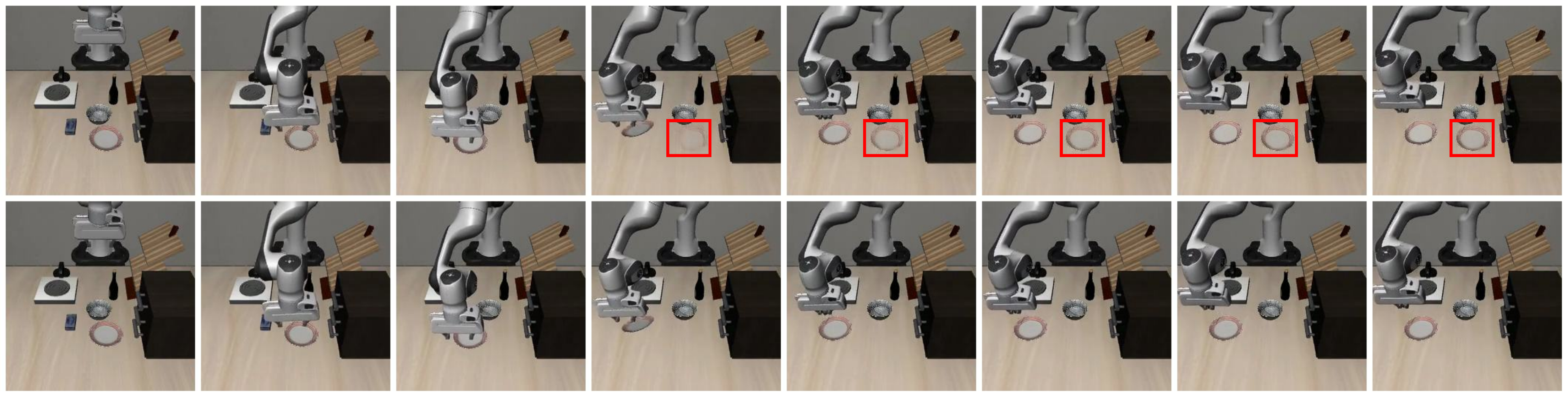}
        \label{fig:compare_ghost_1}
    \end{subfigure}
    \vspace{4pt}
    \begin{subfigure}{\textwidth}
        \centering
        \includegraphics[width=\textwidth]{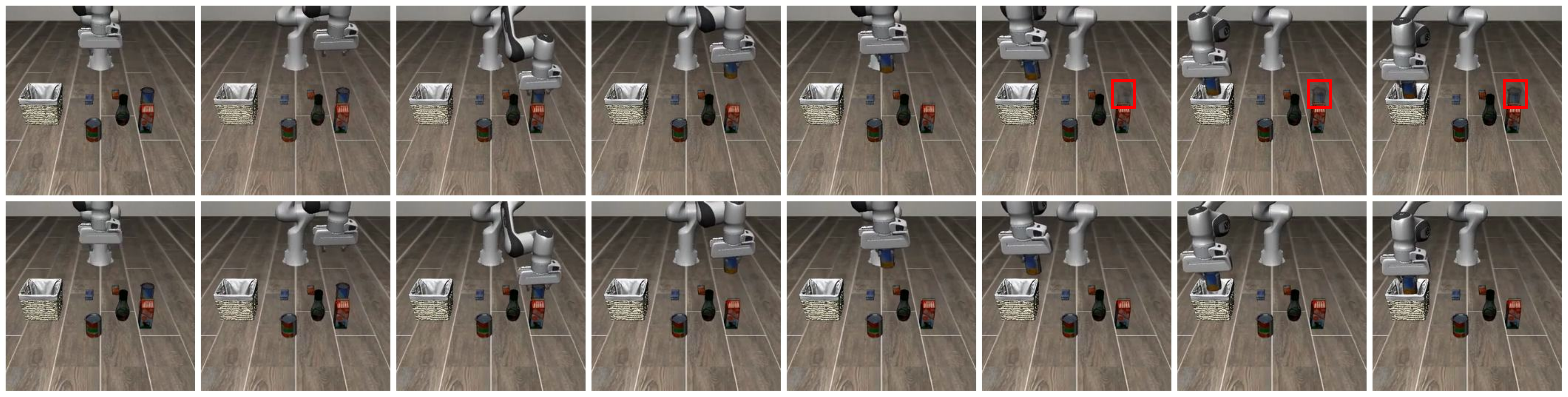}
        \label{fig:compare_ghost_2}
    \end{subfigure}
    \vspace{4pt}
    \begin{subfigure}{\textwidth}
        \centering
        \includegraphics[width=\textwidth]{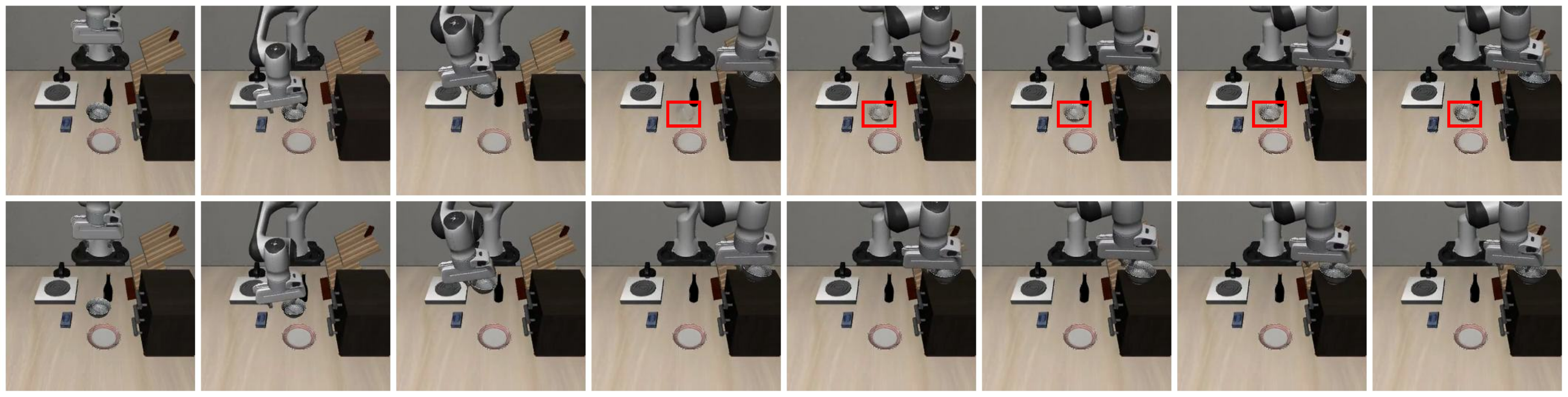}
        \label{fig:compare_ghost_3}
    \end{subfigure}
    \caption{\textbf{Qualitative examples of first-frame ghosting and its mitigation via progressive first-frame timestep noise.}
    For each task, we show two world-model rollouts produced from the same initial observation and the same action sequence, differing only in whether progressive first-frame timestep noise is applied at inference.
    \textbf{Top row} of each subfigure: rollout \emph{without} progressive first-frame timestep noise. The model over-anchors to the first frame $\mathbf{z}^{(0)}$ and copies stale early-frame content into later autoregressive steps, producing visible \textbf{ghost artifacts} (red boxes)---e.g., an object that has already been moved or grasped reappears at its original location.
    \textbf{Bottom row}: rollout \emph{with} progressive first-frame timestep noise applied. The anchor's influence is gradually attenuated as the autoregressive horizon extends, so the model relies on recent context to unroll its learned physical dynamics, eliminating ghosting while still preserving global scene consistency.}
    \label{fig:compare_ghost}
\end{figure}

We provide qualitative examples and analysis on the first-frame ghosting problem in Figure~\ref{fig:compare_ghost}, showcasing the efficacy of the progressive injected first-frame noise timestep  during inference.

\subsection{Qualitative WM Rollout results on Real-World Experiments}
\label{sec:appendix_realworld_rollout}
\begin{figure}[t]
    \centering
    \begin{subfigure}{\textwidth}
        \centering
        \includegraphics[width=\textwidth]{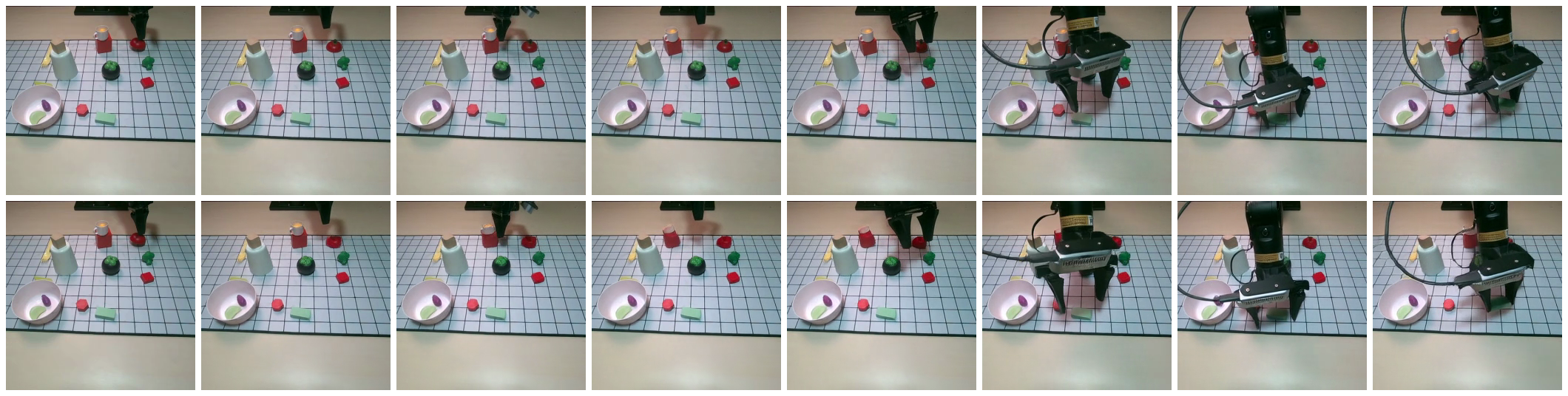}
        \label{fig:rollout_real_1}
    \end{subfigure}
    \vspace{4pt}
    \begin{subfigure}{\textwidth}
        \centering
        \includegraphics[width=\textwidth]{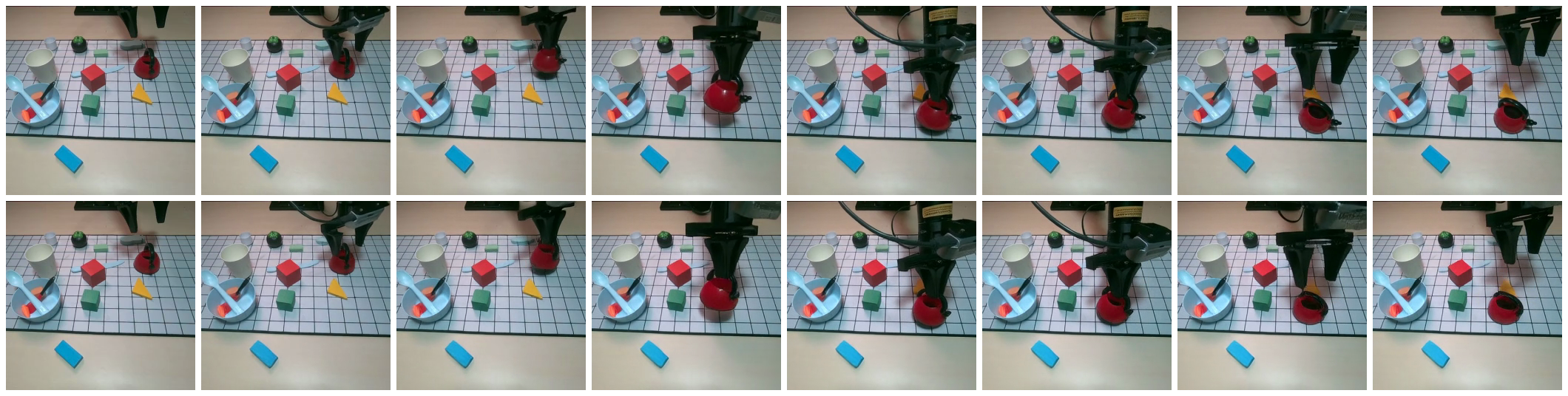}
        \label{fig:rollout_real_2}
    \end{subfigure}
    \vspace{4pt}
    \begin{subfigure}{\textwidth}
        \centering
        \includegraphics[width=\textwidth]{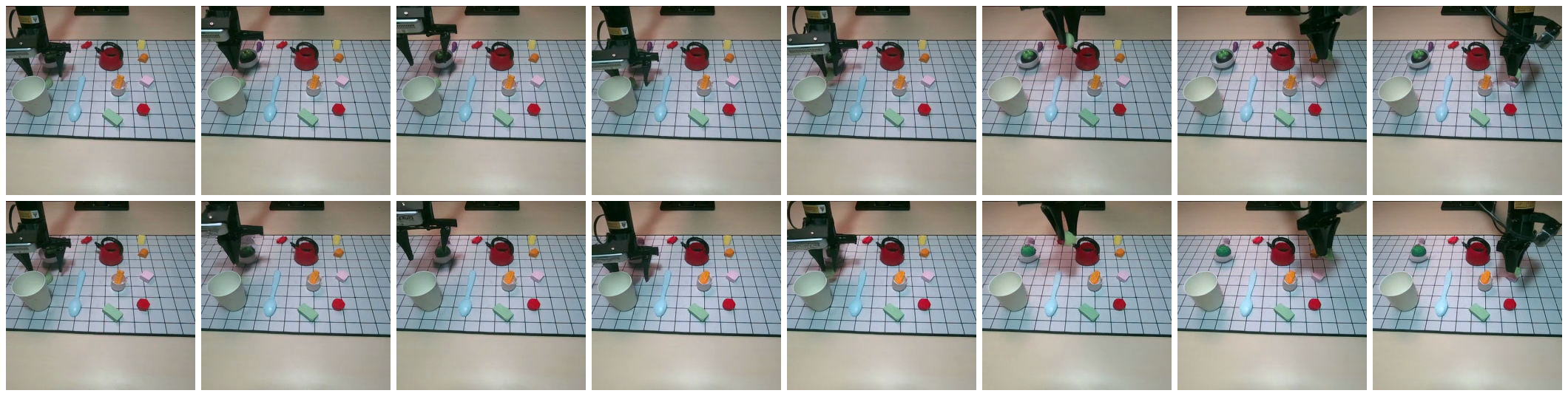}
        \label{fig:rollout_real_3}
    \end{subfigure}
    \vspace{4pt}
    \begin{subfigure}{\textwidth}
        \centering
        \includegraphics[width=\textwidth]{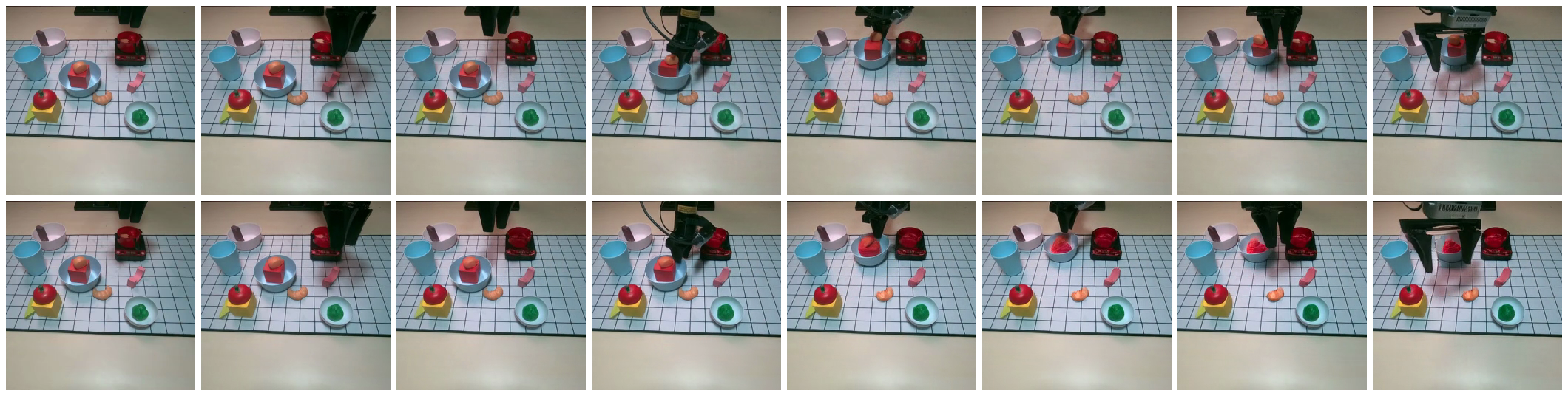}
        \label{fig:rollout_real_4}
    \end{subfigure}
    \vspace{4pt}
    \begin{subfigure}{\textwidth}
        \centering
        \includegraphics[width=\textwidth]{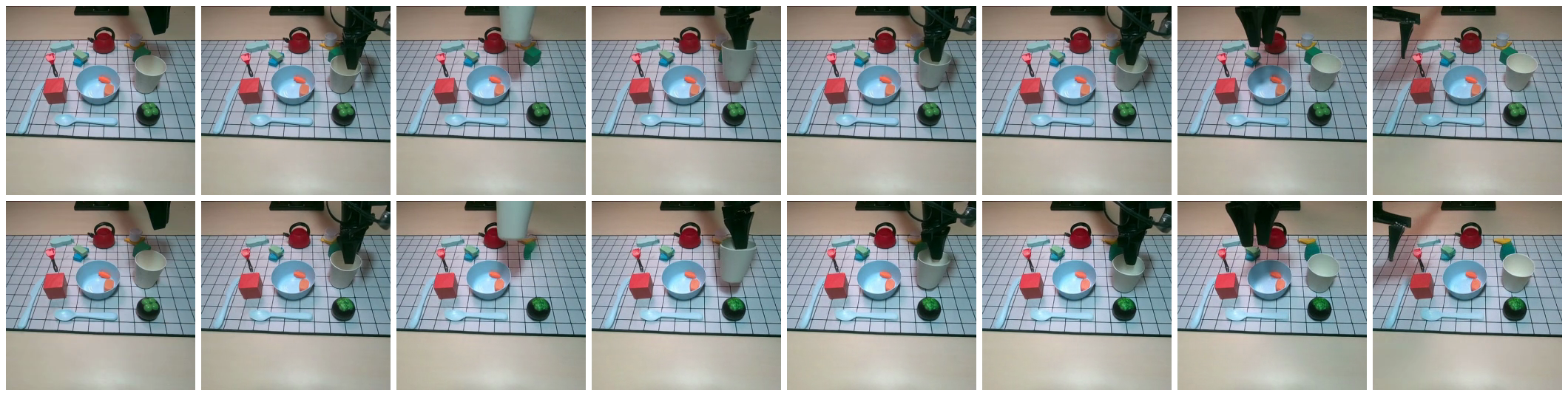}
        \label{fig:rollout_real_5}
    \end{subfigure}
    \caption{\textbf{Qualitative real-world rollouts of our task-agnostic world model.}
    \textbf{Top row} of each subfigure: the ground-truth real-world video executed on the AgileX Piper arm.
    \textbf{Bottom row}: the corresponding autoregressive prediction from our WM, conditioned on the same initial observation $\mathbf{o}_0$ and the same teleoperated action sequence. These results are evaluated on entirely unseen scene layouts absent from the WM's play-data training set.
    }
    \label{fig:rollout_real}
\end{figure}

We provide qualitative WM rollout results in Figure~\ref{fig:rollout_real}, evaluated on entirely unseen scene layouts absent from the WM's play-data training set and without any fine-tuning. Despite being pre-trained only on uncurated, task-agnostic play data --- with \emph{zero} exposure to these downstream tasks or scene layouts --- the WM tracks gripper trajectory, contact events, and object dynamics over horizons of 121--201 frames, producing rollouts faithful enough to drive RL post-training entirely in imagination.

\subsection{Additional Results and Analysis on Dual-Noise Verification}
\label{sec:appendix_ghost_analysis}
\textbf{Effect of DNV on Early Training.} While Table~\ref{tab:ablation} in the main text reports the final RL performance, 
Table~\ref{tab:dualnoise_ablation_20} further reveals that DNV accelerates the early phase of training.
At just 10--20 RL steps, DNV already delivers sizable gains over the no-DNV baseline on most suites,
indicating that filtering hallucinated false positives from the outset 
provides cleaner gradient signal and enables faster initial policy improvement.

\begin{figure}[t]
    \centering
    \begin{subfigure}{\textwidth}
        \centering
        \includegraphics[width=\textwidth]{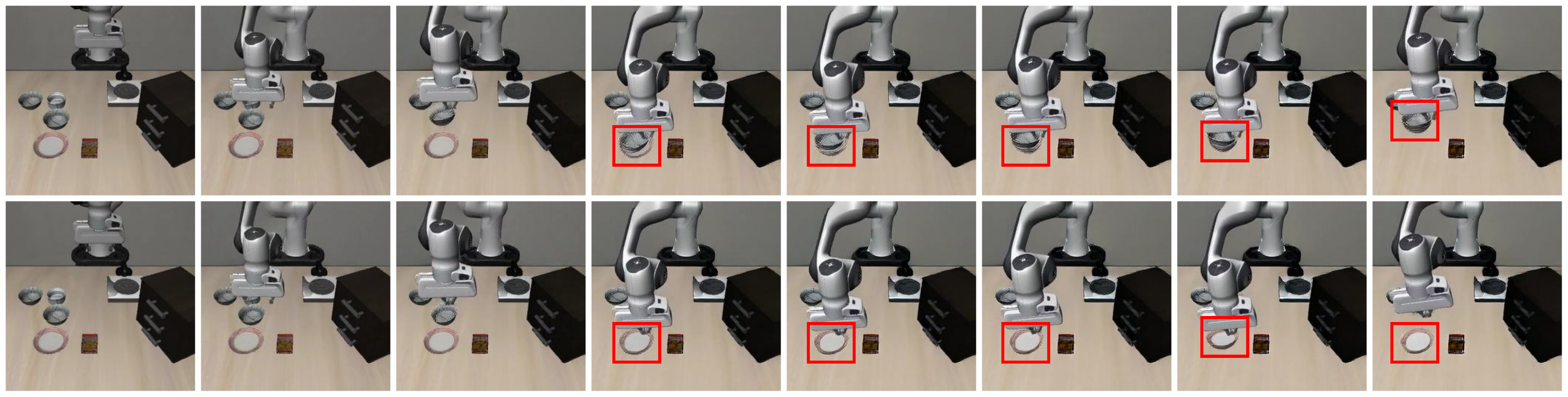}
        \caption{Task: \emph{``pick up the black bowl between the plate and the ramekin and place it on the plate''} from LIBERO-Spatial.}
        \label{fig:compare_dnv_1}
    \end{subfigure}
    \vspace{4pt}
    \begin{subfigure}{\textwidth}
        \centering
        \includegraphics[width=\textwidth]{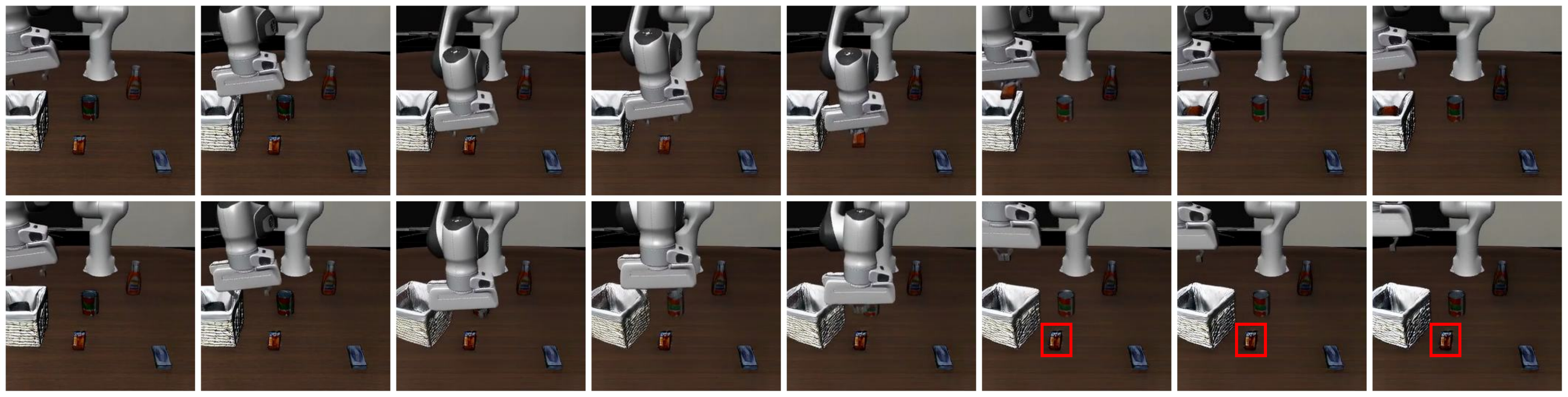}
        \caption{Task: \emph{``put both the cream cheese box and the butter in the basket''} from LIBERO-Long.}
        \label{fig:compare_dnv_2}
    \end{subfigure}
    \vspace{4pt}
    \begin{subfigure}{\textwidth}
        \centering
        \includegraphics[width=\textwidth]{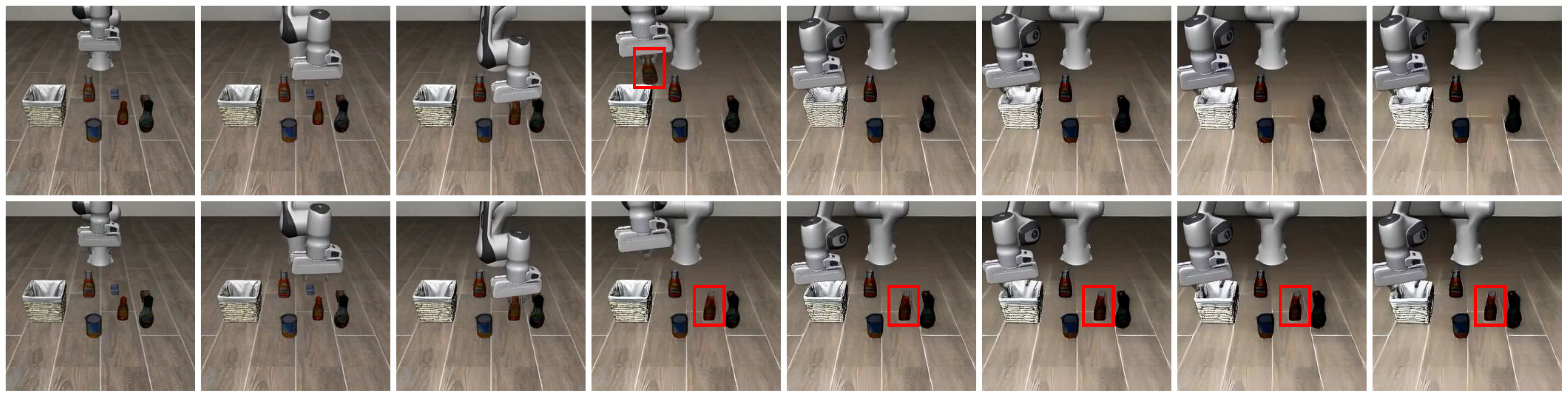}
        \caption{Task: \emph{``pick up the bbq sauce and place it in the basket''} from LIBERO-Object. }
        \label{fig:compare_dnv_3}
    \end{subfigure}
    \caption{\textbf{Qualitative examples of Dual-Noise Verification (DNV).}
    For each task, we show two world-model rollouts produced under the \emph{same action sequence} but with \emph{independently re-sampled initial diffusion noise} at every autoregressive step.
    \textbf{Top row} of each subfigure: the original imagined rollout, on which the VLM reward returns a \textbf{success} verdict.
    \textbf{Bottom row}: the second-pass rollout using the same action sequence, under fresh noise, on which the VLM returns a \textbf{failure} verdict.
    The disagreement reveals that the apparent first-pass success was a hallucination unsupported by the WM's true predictive distribution rather than a faithful physical outcome.
    DNV detects such hallucination and excludes these rollouts from the GRPO update, curbing false-positive-driven reward hacking on unseen tasks.}
    \label{fig:compare_dnv}
\end{figure}

\textbf{Qualitative DNV Examples.}
Figure~\ref{fig:compare_dnv} visualizes how DNV detects hallucinated successes: 
replaying the same actions under different diffusion noise in the WM may produces divergent outcomes, 
and the VLM reward disagreement directly exposes rollouts whose apparent success is an artifact of the WM generation stochasticity rather than faithful physical dynamics.

\textbf{DNV Computational Overhead.}
DNV requires a second-pass WM generation only for rollouts initially judged as successes by the VLM,
which typically constitute a minority of the GRPO group.
In practice, this adds approximately $1.3\times$ wall-clock overhead to the WM inference stage per RL step,
a modest cost given that DNV yields consistent gains,
especially on long-horizon tasks where hallucinations compound most severely (LIBERO-Long).

\begin{table}[t]
  \caption{\textbf{Additional Results on dual-noise verification.}
  Reward is fixed to VLM zero-shot (Qwen3-VL).
  We report success rate (\%) at two early RL training checkpoints (step 10 and step 20) to show DNV's positive effect on training at early stage.}
  \label{tab:dualnoise_ablation_20}
  \centering
  \setlength{\tabcolsep}{2.5pt}
  \small
  \resizebox{\textwidth}{!}{%
  \begin{tabular}{lc cc cc cc cc}
    \toprule
    & & \multicolumn{2}{c}{Spatial} & \multicolumn{2}{c}{Object} & \multicolumn{2}{c}{Goal} & \multicolumn{2}{c}{Long} \\
    \cmidrule(lr){3-4} \cmidrule(lr){5-6} \cmidrule(lr){7-8} \cmidrule(lr){9-10}
    WM & DNV & @10 & @20 & @10 & @20 & @10 & @20 & @10 & @20 \\
    \midrule
    \multirow{2}{*}{Zero-Shot}
    & w/o     & 55.6{\scriptsize(+1.0)} & 57.8\,{\scriptsize(+3.2)} & 53.2\,{\scriptsize(+6.8)} & 46.0\,{\scriptsize(-0.4)} & 54.4\,{\scriptsize(+2.2)} & 54.2\,{\scriptsize(+2.0)} & 21.2\,{\scriptsize(+1.0)} & 26.8\,{\scriptsize(+6.6)} \\
    & w/   & 55.8{\scriptsize(+1.2)} & 61.4\,{\scriptsize(+6.8)} & 49.0\,{\scriptsize(+2.6)} & 47.2\,{\scriptsize(+0.8)} & 55.2\,{\scriptsize(+3.0)} & 58.6\,{\scriptsize(+6.4)} & 21.0\,{\scriptsize(+0.8)} & 26.8\,{\scriptsize(+6.6)} \\
    \midrule
    \multirow{2}{*}{Co-Train}
    & w/o     & 59.8{\scriptsize(+5.2)} & 67.2\,{\scriptsize(+12.6)} & 56.2\,{\scriptsize(+9.8)} & 56.4\,{\scriptsize(+10.0)} & 56.2\,{\scriptsize(+4.0)} & 57.0\,{\scriptsize(+4.8)} & 22.2\,{\scriptsize(+2.0)} & 26.4\,{\scriptsize(+6.2)} \\
    & w/    & 62.0\,{\scriptsize(+7.4)} & 66.6\,{\scriptsize(+12.0)} & 58.0\,{\scriptsize(+11.6)} & 60.2\,{\scriptsize(+13.8)} & 58.4\,{\scriptsize(+6.2)} & 59.6\,{\scriptsize(+7.4)} & 25.6\,{\scriptsize(+5.4)} & 27.0\,{\scriptsize(+6.8)} \\
    \bottomrule
  \end{tabular}
  }
\end{table}

\end{document}